\documentclass[10pt,twocolumn,letterpaper]{article}

\usepackage{cvpr}
\usepackage{times}
\usepackage{epsfig}
\usepackage{graphicx}
\usepackage{amsmath}
\usepackage{amssymb}
\usepackage{multirow}
\usepackage{hhline}
\usepackage{caption}
\usepackage{subcaption}

\newcommand{\eat}[1]{{}}

\usepackage{amsmath,amssymb,amsopn}
\usepackage{bm} 
\usepackage{mathtools}

\newcommand{\vct}[1]{\boldsymbol{#1}} 
\newcommand{\mat}[1]{\boldsymbol{#1}} 
\newcommand{\vtheta}{\vct{\theta}}

\newcommand{\vc}{\vct{c}}

\newcommand{\vf}{\vct{f}}

\newcommand{\vl}{\vct{l}}

\newcommand{\vn}{\vct{n}}

\newcommand{\vp}{\vct{p}}

\newcommand{\vt}{\vct{t}}

\newcommand{\vw}{\vct{w}}
\newcommand{\vx}{\vct{x}}
\newcommand{\vy}{\vct{y}}

\newcommand{\field}[1]{\mathbb{#1}}
\newcommand{\R}{\field{R}} 

\newcommand{\Map}[1]{\mathcal{#1}}

\newcommand{\calS}{\Map{S}}

\newcommand{\calU}{\Map{U}}

\newcommand{\cst}[1]{\mathsf{#1}}

\newcommand{\cD}{\cst{D}}

\newcommand{\cM}{\cst{M}}
\newcommand{\cN}{\cst{N}}

\newcommand{\cU}{\cst{U}}

\newcommand{\inner}[2]{\left\langle#1, #2\right\rangle}

\DeclareMathOperator{\argmin}{arg\,min}
\newcommand{\defeq}{\vcentcolon=}

\newcommand{\specialcell}[2][c]{%
  \begin{tabular}[#1]{@{}c@{}}#2\end{tabular}}
  
\newcommand{\pd}{{\vw}}
\newcommand{\irYm}{\overline{Y_m}}

\newcommand{\ours}{{Fast\textbf{0}Tag}}

\usepackage[pagebackref=true,breaklinks=true,letterpaper=true,colorlinks,bookmarks=false]{hyperref}

\cvprfinalcopy 

\def\cvprPaperID{464} 

\ifcvprfinal\fi
\begin{document}

\title{ Fast Zero-Shot Image Tagging}

\author{Yang Zhang, Boqing Gong, and Mubarak Shah\\ 
Center for Research in Computer Vision, University of Central Florida, Orlando, FL 32816\\
{\tt\small yangzhang@knights.ucf.edu, bgong@crcv.ucf.edu, shah@crcv.ucf.edu}
}

\maketitle

\begin{abstract}

The well-known word analogy experiments show that the recent word vectors capture fine-grained linguistic regularities in words by linear vector offsets, but it is unclear how well the simple vector offsets can encode visual regularities over words. We study a particular image-word relevance relation in this paper. Our results show that the word vectors of   relevant tags for a given image   rank ahead of the irrelevant tags, along a principal direction in the word vector space. Inspired by this observation, we propose to solve image tagging by  estimating the principal direction for an image. Particularly, we exploit linear mappings and nonlinear deep neural networks to approximate the principal direction from an input  image. We arrive at  a quite versatile tagging model. It runs fast given a test image, in constant time w.r.t.\ the training set size. It not only gives superior performance for the conventional tagging task on the NUS-WIDE dataset, but also outperforms  competitive baselines on annotating images with previously {\bf unseen} tags. 

\eat{
Ranking labels that are not seen during training for an image, know as zero-shot tagging, is receiving more and more attention. We have proposed Fast0Tag which is able to rank both seen and unseen labels for query image. Fast0Tag is a neural network transforming each query image to a unique linear ranking hyperplane in embedding space, so that rank hyperplane could ranks labels, regardless they have appeared during training or not, in the embedding space for the query. We have validated presumptions and discussed the optimization of neural network. We have compared Fast0Tag with state-of-art methods as well as some proposed baselines in both conventional and zero-shot labeling task on NUS tagging dataset. Fast0Tag has shown very promising performance improvement compared with baselines.
}

\end{abstract}

\section{Introduction}
\label{sIntro}

\begin{figure}[t]
\begin{center}
\includegraphics[width=0.9\linewidth]{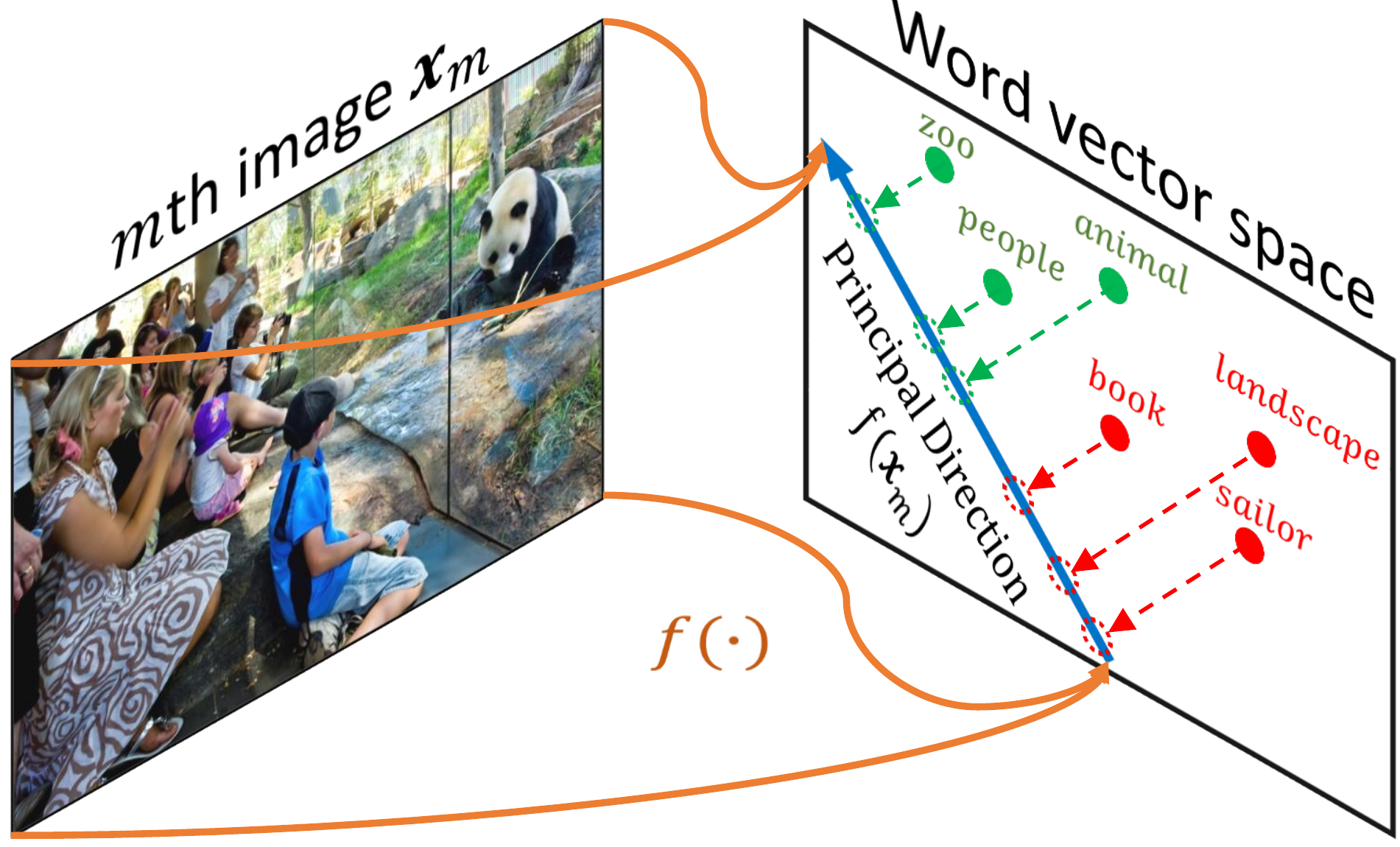}
\end{center}
   \caption{Given an image, its relevant tags\rq{} word vectors  rank ahead of  the irrelevant tags\rq{} along some direction in the word vector space. We call that direction the \textbf{principal direction} for the image. To solve the problem of image tagging, we thus learn a function $\vf(\cdot)$ to approximate the principal direction from an image. This function takes as the input an image $\vx_m$ and outputs a vector $\vf(\vx_m)$ for defining the principal direction in the word vector space. }
\label{fConcept}
\vspace{-10pt} 
\end{figure}

Recent advances in the vector-space representations of words~\cite{mikolov_efficient_2013,mikolov_distributed_2013,pennington_glove:_2014} have benefited
both NLP~\cite{socher_parsing_2013,zou_bilingual_2013,tellex_quantitative_2003} and  computer vision tasks such as zeros-shot learning~\cite{socher_zero-shot_2013,frome_devise:_2013,akata_label-embedding_2013} and image captioning~\cite{lebret_phrase-based_2015-1,karpathy_deep_2014,kiros_unifying_2014}. The use of word vectors in NLP is grounded on  the fact that the {fine-grained} \textbf{linguistic} regularities over words are captured by linear word vector offsets---a key observation from the well-known word analogy experiments~\cite{mikolov_linguistic_2013,pennington_glove:_2014}, such as the syntactic relation $dance-dancing\approx fly-flying$ and semantic relation ${king}-{man}\approx{queen}-{woman}$. However, it is unclear whether the \textbf{visual} regularities over words, which are implicitly used in the aforementioned computer vision problems, can still be  encoded by the simple  vector offsets.



In this paper, we are interested in the problem of image tagging, where an image (e.g., of a zoo in Figure~\ref{fConcept}) calls for a partition of a vocabulary of words into two disjoint sets according to the image-word relevance (e.g., relevant tags $Y=\{{people,animal,zoo}\}$ and irrelevant ones $\overline{Y}=\{{sailor,book,landscape}\}$). This partitioning of words, $(Y,\overline{Y})$, is essentially different from the {fine-grained} syntactic (e.g., {dance} to {dancing}) or semantic (e.g., {king} to {man}) relation tested in the word analogy experiments. Instead, it is about the relationship between two {sets} of words due to a visual image. Such a relation in words is semantic and descriptive, and  focuses on \textbf{visual association}, albeit relatively {coarser}.  In this case, do the  word vectors still offer the nice property, that {the simple linear vector offsets can depict the visual (image) association relations in words?} For the example of the zoo, while humans are capable of easily answering that the words in ${Y}$ are more related to the zoo than those in ${\overline{Y}}$, can such zoo-association relation in words be expressed by the 9 pairwise word vector offsets $\{{people}-{sailor}, {people}-{book}, \cdots, {zoo}-{landscape}\}$ between the relevant ${Y}$ and irrelevant ${\overline{Y}}$ tags' vectors?

\begin{figure*}
\centering
\vspace{-12pt}
\includegraphics[width=0.8\linewidth]{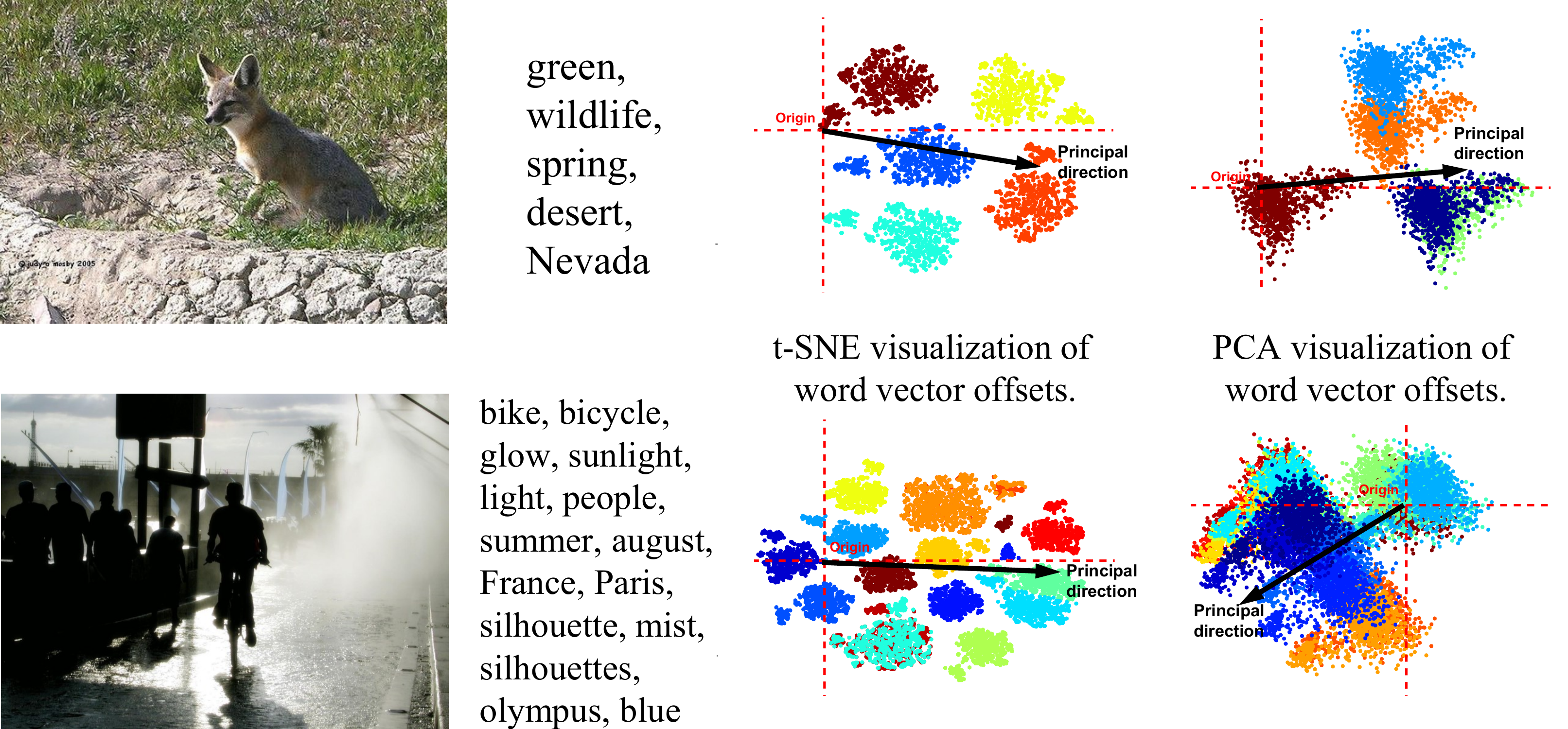}
   \caption{Visualization  of the offsets between relevant tags\rq{} word vectors and irrelevant ones\rq{}. \emph{Note that each vector from the origin to a point is an offset between two word vectors.}
   The relevant tags are shown beside the images~\cite{chua_nus-wide:_2009}. }
\label{fVisualize}
\vspace{-12pt}
\end{figure*}

One of the main contributions of this paper is to empirically examine the above two questions (cf.\ Section~\ref{sRankability}). Every image introduces a visual association rule  $(Y,\overline{Y})$ over  words. Thanks to the large number of images in  benchmark datasets for image tagging, we are able to examine many distinct \emph{visual association regulations} in words and the corresponding \emph{vector offsets} in the word vector space. Our results reveal a somehow surprising connection between the two:  the offsets between the vectors of the relevant tags $Y$ and those of the irrelevant $\overline{Y}$ are along about the same direction, which we call the \textbf{principal direction}. See Figure~\ref{fVisualize} for the visualization of some vector offsets. In other words, there exists at least one vector (direction) $\vw$ in the word vector space, such that its inner products with  the vector offsets between ${Y}$ and ${\overline{Y}}$ are greater than 0, i.e., $\forall \vp\in{Y}$, $\forall \vn\in{\overline{Y}}$, 
\begin{align}
\inner{\vw}{\vp-\vn} > 0 \; \text{equivalently,}\; \inner{\vw}{\vp} > \inner{\vw}{\vn}, \label{eFinding}
\end{align}
where the latter reads that the vector $\vw$ \emph{ranks} all relevant words  ${Y}$ (e.g., for the zoo image) ahead of the irrelevant ones  $\overline{Y}$. For brevity, we overload the notations $Y$ and $\overline{Y}$ to respectively denote the vectors of the words in them. 

The visual association relations in words thus represent themselves by the {(linear)} rank-abilities of the corresponding word vectors.  This result reinforces the conclusion from the word analogy experiments that, for a
single word multiple relations are embedded  in the high dimensional
space~\cite{mikolov_linguistic_2013,pennington_glove:_2014}. Furthermore, those relations can be expressed by simple linear vector arithmetic. 


Inspired by the above observation, we propose to solve the image tagging problem by estimating the principal direction, along which the relevant tags rank ahead of the irrelevant ones in the word vector space. Particularly, we exploit linear mappings and  deep neural networks to approximate the principal direction from each input  image. This is a grand new point of view to image tagging and results in a quite versatile tagging model. It operates fast given a test image, in constant time with respect to the training set size. It not only gives superior performance for the conventional tagging task, but  is also capable of assigning  novel tags from an open vocabulary, which are unseen at the training stage. We do not assume any \textit{a priori} knowledge about these unseen tags as long as they are in the same vector space as the seen tags for training. To this end, we name our approach fast zero-shot image tagging (Fast\textbf{0}Tag) to recognize that it possesses the advantages of both FastTag~\cite{chen_fast_2013} and zero-shot learning~\cite{lampert_attribute-based_2014,fu_transductive_2014,fu_transductive_2015}.


In sharp contrast to our approach, previous image tagging methods can only annotate test images with the  tags seen at training except~\cite{fu_transductive_2015}, to the best of our knowledge. Limited by the static and usually small number of seen tags in the training data, these models are frequently challenged in practice. For instance, there are about 53M tags on Flickr and the number is rapidly growing. \eat{See Section~\ref{sRelated} for more discussion.} The work of~\cite{fu_transductive_2015} is perhaps the first attempt to generalize an image tagging model to unseen tags. Compared to the proposed method, it depends on two extra assumptions. One is that the unseen tags are known \textit{a priori} in order to tune the model towards their combinations. The other is that the test images are known \textit{a priori}, to regularize the model. Furthermore, the generalization of~\cite{fu_transductive_2015} is limited to a very small number, $\cU$, of unseen tags, as it has to consider all the $2^\cU$ possible combinations.

To summarize, our first main contribution is on the analyses of the visual association relations in  words due to images, and how they are captured by word vector offsets. We hypothesize and empirically verify that, for each visual association rule $(Y,\overline{Y})$, in the word vector space there exists a principal direction, along which the relevant words\rq{} vectors rank ahead of the others\rq{}. Built upon this finding, the second contribution is a novel image tagging model, Fast\textbf{0}Tag, which is fast and generalizes to open-vocabulary unseen tags. Last but not least, we explore three different image tagging scenarios: \textbf{conventional} tagging which assigns seen tags to images, \textbf{zero-shot} tagging which annotates images by (a large number of) unseen tags, and \textbf{seen/unseen} tagging which tags images with both seen and unseen tags. In contrast, the existing work tackles either conventional tagging, or zero-shot tagging with very few unseen tags. Our Fast\textbf{0}Tag gives superior results over  competitive baselines under all the three testing scenarios.

\eat{
Image tagging and tags retrieval is always a popular in computer vision community. The emergence of large scale images and tags drive us to develop more powerful image annotation tools. Unfortunately, the image tagging methods are not able to reason between semantic labels. For example, they not able to retrieve the tag "hill" even there are images labeled with "mountain" in the training set. The reason behind that is they are considering each label as discrete, equivalent item rather than a "word". The only thing they care about is how to correctly retrieve those items appeared in the training set, which are very limited. As number of tags are increasing and 

In order to address this probem, we need to build a uniform framework for both semantic and visual domain. Recently developed zero-shot learning method, which classify objects into unseen classes, are innovative and it made the first concrete step to bridge the gap between semantic and visual domain, but still it is incomplete because it only models the top one label by derive a simple projection between embedding space and visual space. As we are modeling an image and rank of a set of labels, which is a more complicated problem and is not a projection problem.

Following the discover revealed in (cite something) shown in figure.~\ref{fig:intuition}. Given the fact that vector directions between different words encode certain order properties, we developed an intuition that we could model the word vector space with rank functions which determine the belonging of tags to a image.

Thus we proposed Fast0Tag as shown in figure.~\ref{fig:overview}, a zero-shot label ranking approach jointly optimizing both visual and semantic modalities by mapping each unique image to an exclusive linear ranking hyperplane in the semantic space. We optimized Fast0Tag $f(\cdot)$ so that the mapped hyperplane could properly ranks tags, regardless it is seen or not during training, in the semantic space for this image.

In this paper we made following contributions: 1. We proved that the tags in embedding space is predictable though trained rank hyperplane. 2. We proposed Fast0Tag to estimate rank hyperplane directly from image to address the zero-shot ranking problem. It outperformed state-of-art methods and proposed baselines. 3. We proposed some baselines. Although they did not outperform the proposed method, they offer ideas how to address zero-shot ranking problem from different perspectives.
}

\section{Related work}
\label{sRelated}

\paragraph{Image tagging.} Image tagging aims to assign relevant tags  to an image or to return a ranking list of tags. In the literature this problem has been mainly approached from the tag ranking perspective. In the generative methods, which involve topic models~\cite{barnard_matching_2003,monay_plsa-based_2004,yakhnenko_annotating_2008,niu_semi-supervised_2014} and mixture models~\cite{lavrenko_model_2003,jeon_automatic_2003,tariq_feature-independent_2015,feng_multiple_2004,carneiro_supervised_2007,dehghan_improving_2014}, the candidate tags are naturally ranked according to their probabilities conditioned on the test image. For the non-parametric nearest neighbor based methods~\cite{makadia_baselines_2010,mei_coherent_2008,li_learning_2009,kalayeh_nmf-knn:_2014,guillaumin_tagprop:_2009,lee_visually_2014,zhu_adaptive_2014}, the tags for the test image are often ranked by the votes from some training images. The nearest neighbor based algorithms, in general, outperform those depending on generative models~\cite{kalayeh_nmf-knn:_2014,li_socializing_2015}, but suffer from high computation costs in both training and testing. The recent FastTag algorithm~\cite{chen_fast_2013} is magnitude faster and achieves comparable results with the nearest neighbor based methods. Our Fast\textbf{0}Tag shares the same level of low complexity as FastTag. The embedding method~\cite{weston_large_2010} assigns ranking scores to the tags by a  cross-modality mapping between images and tags. This idea is further exploited using deep neural networks~\cite{gong_deep_2013}. Interestingly, none of these methods learn their models explicitly for the ranking purpose except~\cite{weston_large_2010,gong_deep_2013}, although they all rank the candidate tags for the test images. Thus, there exists a mismatch between the models learned and the actual usage of the models, violating the principle of Occam's razor. We use a ranking loss in the same spirit as~\cite{weston_large_2010,gong_deep_2013}. 

In contrast to our Fast\textbf{0}Tag, which can rank both seen and an arbitrary number of unseen tags for test images, the aforementioned approaches only assign tags to images from a closed vocabulary seen at the training stage. An exception is by Fu et al.~\cite{fu_transductive_2014}, where the authors consider pre-fixed $\cU$ unseen tags and learn a multi-label model to account for all the $2^{\cU}$ possible combinations of them. This method is limited to  a small number $\cU$ of unseen tags. 
\vspace{-10pt}

\eat{
Image tagging method builds model which assigns or ranks a tag set to a given unannotated query image. According to ~\cite{li_socializing_2015}, there are three types of methods: instance-based~\cite{sigurbjornsson_flickr_2008}, model-based~\cite{guillaumin_tagprop:_2009,chen_fast_2013} and transductive-based. These methods aims to assign or rank tags that has already been seen in the training set while our ranking model does not require seeing tags to rank them. And there method ranks discrete items while ours ranks continuous representation.

Indeed, image tagging has been approached from the tag ranking perspective for decades. 
}

\paragraph{Word embedding.}
Instead of representing words using the traditional one-hot vectors, word embedding maps each word to a continuous-valued vector, by  learning from primarily the statistics of word co-occurrences. Although there are earlier works on word embedding~\cite{rumelhart_learning_1985,deerwester_indexing_1990}, we point out that our work focuses on  the most recent GloVe~\cite{pennington_glove:_2014} and word2vec vectors~\cite{mikolov_linguistic_2013,mikolov_distributed_2013,mikolov_efficient_2013}. As shown in the well-known word analogy experiments~\cite{mikolov_linguistic_2013,pennington_glove:_2014}, both types of word vectors are able to capture fine-grained semantic and syntactic regularities using vector offsets. In this paper, we further show that the simple linear offsets also depict the relatively coarser visual association relations in words.
\vspace{-10pt}
 
\paragraph{Zero-shot learning.}
Zero-shot learning is often used exchange-ably with zero-shot classification, whereas the latter is a special case of the former. Unlike weakly-supervised learning~\cite{moxley_video_2010,fu_relaxing_2015} which learn new concepts by mining noisy new samples, zero-shot classification learns classifiers from seen classes and aims to classify the objects of unseen classes~\cite{palatucci_zero-shot_2009,norouzi_zero-shot_2013,lampert_attribute-based_2014,akata_label-embedding_2013,fu_transductive_2014,jayaraman_zero-shot_2014,norouzi_zero-shot_2013,palatucci_zero-shot_2009,socher_zero-shot_2013}. \eat{It has been considered a promising technique for handling the scarce training data of rare classes.} Attributes~\cite{lampert_learning_2009,farhadi_describing_2009} and word vectors are two of the main semantic sources making zero-shot classification feasible. 

Our Fast\textbf{0}Tag along with~\cite{fu_transductive_2015} enriches the family of zero-shot learning by  zero-shot multi-label classification~\cite{tsoumakas_multi-label_2006}. Fu et al.~\cite{fu_transductive_2015} reduce the problem to zero-shot classification by treating every combination of the multiple labels as a class. We instead directly model the labels and are able to assign/rank many unseen tags for an image.


\eat{
 In order to do this, it is necessary to find a way to transfer knowledge from seen classes to unseen classes. And this is done by using the label embedding of the classes. researchers are trying to find the mapping between image and label embedding so that they could map the query image to the unseen embedding. Please refer to section.~\ref{sDiscussion} for the differences between our method and methods listed here. 

Based on the label embedding they used, there are two major types of zero-shot classification: Attribute-based and word-embedding-based. In attribute base approaches, each class label is described as a visual attributes vector. There are a lot of such method~\cite{lampert_attribute-based_2014,akata_label-embedding_2013,lampert_learning_2009}. The advantage is that attribute is visually distinct embedding which is designed specially for zero-shot learning. And since this embedding qualifies the occurrence of attributes, researcher can also model attributes as well~\cite{jayaraman_zero-shot_2014}. But the disadvantage is also obvious: It require effort to manually describe each class with attribute, hence it would be laboriously expensive when extends this to large dataset.

Another approach is considering word vector as label embedding.  Such method includes ~\cite{norouzi_zero-shot_2013,palatucci_zero-shot_2009,socher_zero-shot_2013,fu_transductive_2015}. Compared with attribute-based approaches, these methods are more general because they could use attributes as input as well. And they free people from annotating label with attributes as well. The drawback is that word vector is often not as representative as attributes.

To our best knowledge, there is only one paper ~\cite{fu_transductive_2015} addressed the problem of zero-shot labeling so far. This paper actually share the same framework with zero-shot classification problem. Instead of find mapping between image and single word vector it estimates mapping between image and synthesized multi-label word vector, which is average summation of word vectors of positive labels. Thus it results exponential computational complex and is inapplicable for medium or larger dataset.
}


\section{The linear rank-ability of word vectors}
\label{sRankability}
Our Fast\textbf{0}Tag approach benefits from the finding that the visual association relation in words, i.e., the partition of a vocabulary of words according to their relevances to an image,  expresses itself in the word vector space as the existence of a principal direction, along which the words/tags relevant to the image rank ahead of the irrelevant ones. This section details the finding.

\eat{
In this paper, we proposed a general framework which learns how to rank items in continuous spaces for a given query in another space and modality. We could incorporate any ranking loss functions into this framework. In this case we provide a way to extend conventional ranking problem from being discrete to being continuous. And we tested this approach by applying it to the problem of newly proposed zero-shot ranking problem, where we need to rank some unseen labels for given query images.
}

\subsection{The regulation over words due to image tagging} 
\label{sRegulation}
We use $\calS$ to denote the seen tags available for training image tagging models and $\calU$ the tags unseen at the training stage.  The training data are in the form of $\{(\vx_m, {Y_m}); m=1,2,\cdots,\cM\}$, where $\vx_m\in\R^{\cD}$ is the feature representation of  image $m$ and ${Y_m}\subset\calS$ are the seen tags relevant to that image. For brevity, we overload the notation ${Y_m}$ to also denote the collection of the corresponding word/tag vectors. 

The \textbf{conventional} image tagging aims to assign seen tags in $\cal{S}$ to the test images. The \textbf{zero-shot} tagging, formalized in~\cite{fu_transductive_2015}, tries to annotate test images using a pre-fixed set of unseen tags $\calU$. In addition to those two scenarios, this paper  considers \textbf{seen/unseen} image tagging, which finds both relevant seen tags from $\calS$ and relevant unseen tags from $\calU$ for the test images. Furthermore, the set of unseen tags $\calU$ could be open and dynamically growing. 

Denote by $\overline{{Y_m}}\defeq\calS\setminus {Y_m}$ the irrelevant seen tags. An image $m$ introduces a visual association regulation to words---the partition $({Y_m}, \overline{{Y_m}})$ of  the seen tags to two disjoint sets. Noting that many fine-grained syntactic and semantic regulations over words can be expressed by  linear word vector offsets, we next examine what properties the vector offsets could offer for this new visual association rule.

\eat{
Before defining zero-shot label ranking problem, which is the problem we would focus in this paper, let us revisit the conventional label ranking problem. Given a set of $\cM_{te}$ instances $\mathbf{x} =\{\vx_1,\vx_2,...,x_{\cM_{te}}\}$ .In our case these are $\cM_{te}$ images. For each $\vx$, we are trying to rank $\cN_{tr}$ labels in another instance set $\mathbf{y} =\{y_1,y_2,...,y_{\cN_{tr}}\}$, so that the correct $y$ ranks higher than incorrect $y$.

In conventional ranking problem, $\mathbf{x}$ are instances represented in continuous space. $\mathbf{y}$ are discrete labels consistent in training set and testing set. A general approach is to estimate a ranking function $f(\vx_m)\rightarrow\vc_m$, where $\vc_m\in\mathbb{R}^{1\times \cN_{tr}}$ quantifies association between the $\cst{m}$th query $\vx_{\cst{m}}$ and $\cN_{tr}$ labels in $\mathbf{y}$.

In the zero-shot label ranking problem, our model are not only ranking $\cN_{tr}$ $\mathbf{y}$ labels presenting in training set, but also ranking additional $\cN_{te}$ unseen tags $\mathbf{y_{unseen}}=\{y_{\cN_{tr}+1},y_{\cN_{tr}+2},...,y_{\cN_{tr}+\cN_{te}}\}$. So we are estimating a ranking function $f(\vx_{\cst{m}},\mathbf{y_{unseen}})\rightarrow\vc_{unseen}$ where $\vc_{unseen}\in\mathbb{R}^{1\times (\cN_{tr}+\cN_{te})}$. Since $\mathbf{y_{unseen}}$ are not presented in the training stage, we need to transfer the knowledge by representing $\mathbf{y_{unseen}}$ in a continuous feature space just like $\mathbf{x}$, known as label embedding space.
}

\subsection{Principal direction and cluster structure} \label{sStructure}

Figure~\ref{fVisualize} visualizes the vector offsets $(\vp-\vn)$, $\forall \vp\in{Y_m}, \forall \vn \in \overline{{Y_m}}$ using t-SNE~\cite{van_der_maaten_visualizing_2008} and PCA for two  visual association rules over words. One is imposed by an image with $5$ relevant tags and the other is with $15$ relevant tags. We observe  two main structures from the vector offsets:
\vspace{-5pt}
\begin{description}
\item[Principal direction.] Mostly, the vector offsets point to about the same direction (relative to the origin), which we call the principal direction, for a given  visual association rule $({Y_m},\overline{{Y_m}})$ in words for image $m$. This implies that the relevant tags ${Y_m}$ rank ahead of the irrelevant ones  $\overline{{Y_m}}$ along the principal direction (cf.\ eq.~(\ref{eFinding})).
\vspace{-15pt}
\item[Cluster structure.] There exist cluster structures in the vector offsets for each visual association regulation over the words. Moreover, all the offsets pointing to the same relevant tag in ${Y_m}$ fall into the same cluster. We differentiate the offsets pointing to different relevant tags by colors in Figure~\ref{fVisualize}.
\vspace{-5pt}
\end{description}
{Can the above two observations generalize? Namely, do they still hold in the high-dimensional word vector space for more visual association rules imposed by other images?} To answer the questions, we next design an experiment to verify the existence of the principal directions in word vector spaces, or equivalently the linear rank-ability of word vectors. We leave the cluster structure for future research.

\subsection{Testing the linear rank-ability hypothesis}
Our experiments in this section are conducted on the validation set (26,844 images, 925 seen tags $\calS$, and 81 unseen tags $\calU$) of NUS-WIDE~\cite{chua_nus-wide:_2009}. The number of relevant seen/unseen tags associated with an image ranges from 1 to 20/117 and on average is 1.7/4.9. See Section~\ref{sExp} for details.

Our objective is to investigate, for any visual association rule $(Y_m, \overline{Y_m})$ in words by image $m$, the existence of the principal direction along which the relevant tags  $Y_m$ rank ahead of the irrelevant tags $\overline{Y_m}$. The proof completes once we find a vector $\vw$ in the word vector space that satisfies the ranking constraints $\inner{\vw}{\vp}>\inner{\vw}{\vn}, \forall \vp\in Y_m, \forall \vn\in\overline{Y_m}$. To this end, we train a linear ranking SVM~\cite{joachims_optimizing_2002} for each visual association rule using all the corresponding pairs $(\vp,\vn)$, then rank the word vectors by the SVM, and finally examine how many constraints are violated. In particular, we employ MiAP, the larger the better (cf.\ Section~\ref{sExp}), to compare the SVM's ranking list with those ranking constraints. We repeat the above process for all the validation images, resulting in 21,863 unique visual association rules.

\paragraph{Implementation of ranking SVM.}
In this paper, we use the implementation of solving ranking SVM in the primal~\cite{chapelle_efficient_2010} with the following formulation:
\[
	\min_{\vw} \;\frac{\lambda}{2}\lVert \vw \rVert^2+\sum_{\vy_i \in Y_m} \sum_{\vy_j\in\irYm} \max(0 , 1-\vw \vy_i+\vw \vy_j)  
\]
where $\lambda$ is the hyper-parameter controlling the trade-off between the objective and the regularization. 

\begin{figure}[h]
\vspace{-5pt}
\centering
\begin{tabular}{cc}
   \hspace*{-0.4cm}\includegraphics[width=0.45\linewidth]{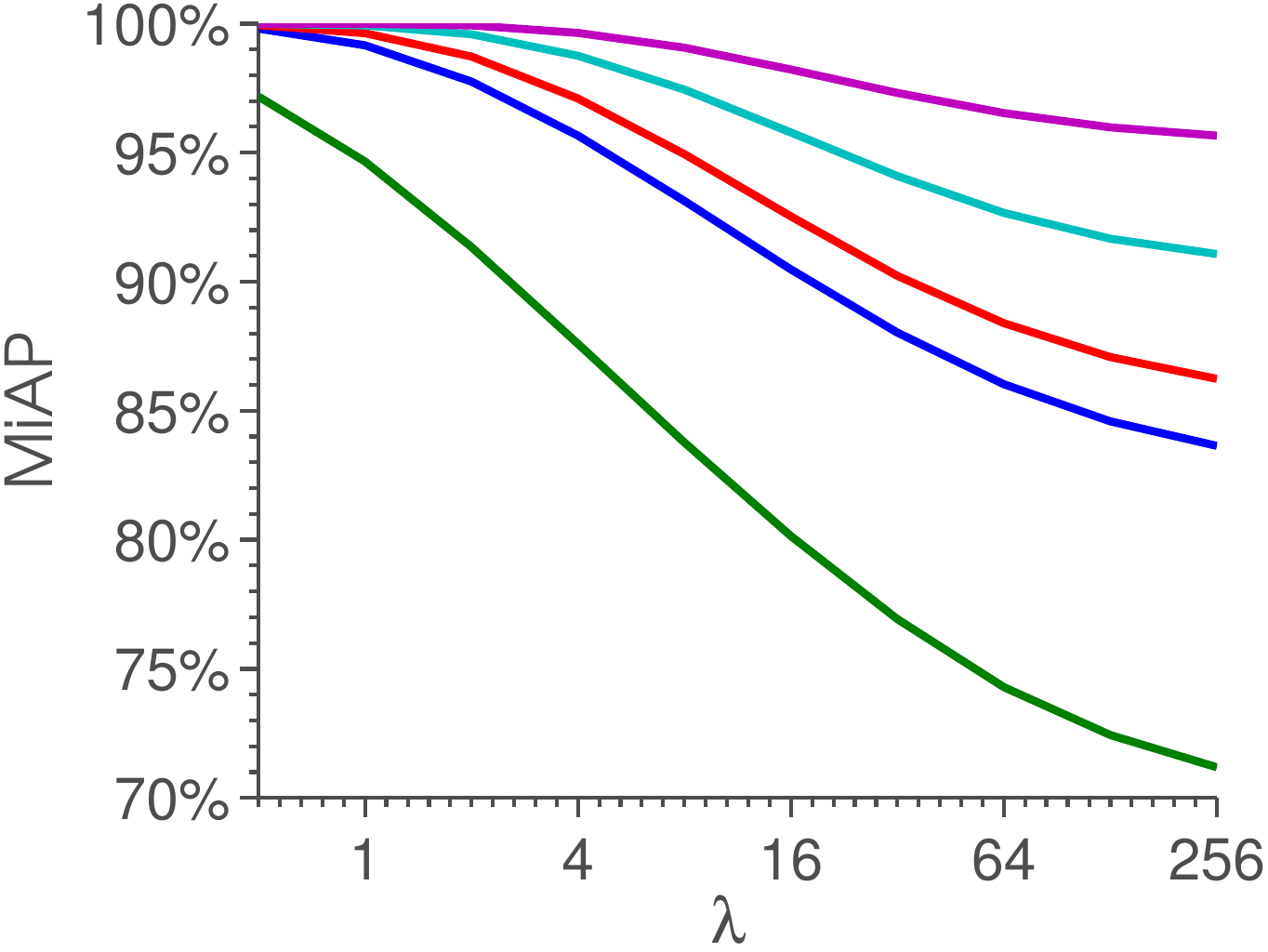} & \hspace*{-0.2cm}\includegraphics[width=0.45\linewidth]{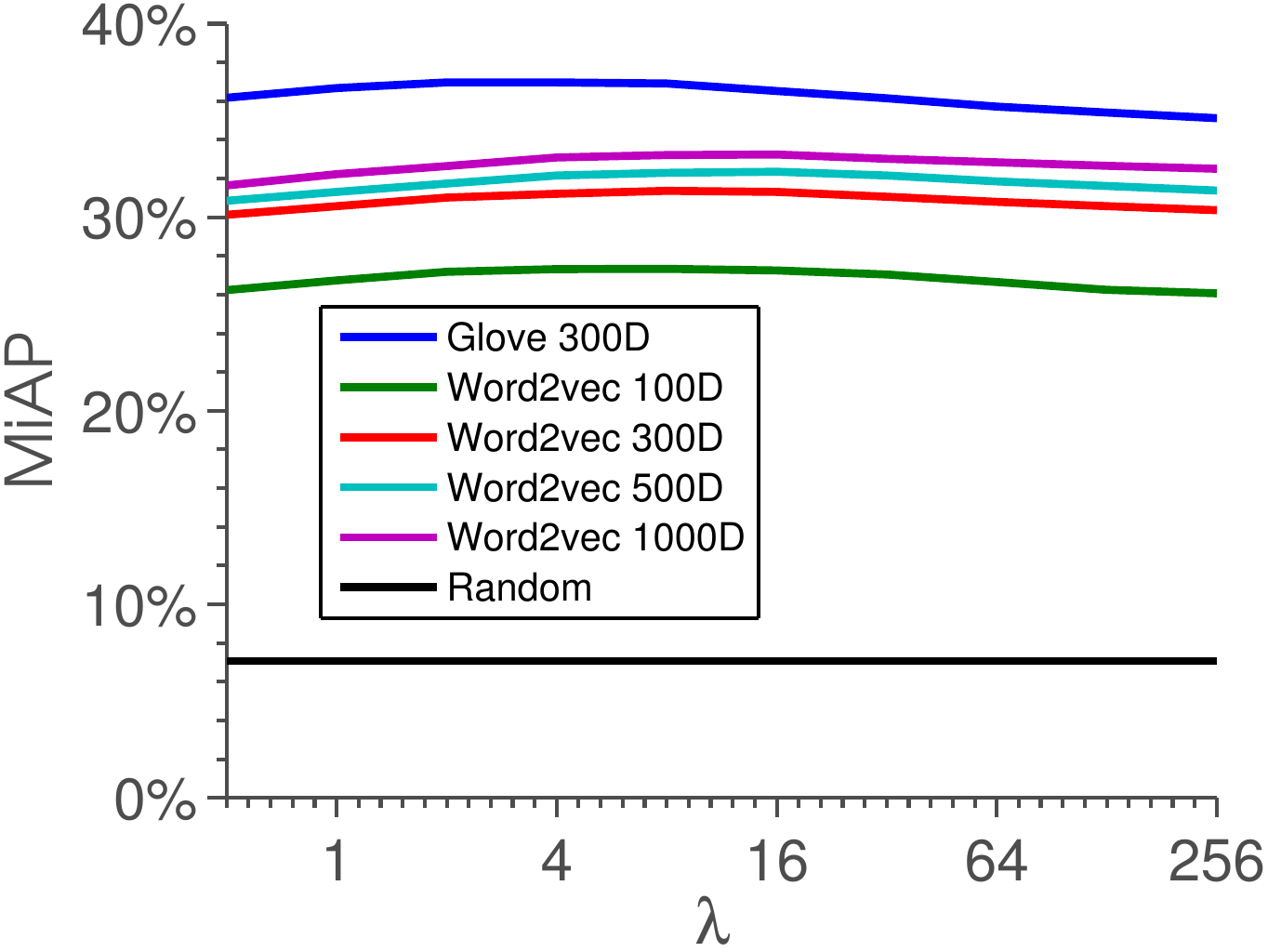} \vspace{-3pt}
\end{tabular}
\vspace{-7pt}
   \caption{The existence (left) and generalization (right) of the principal direction for each visual association rule in words induced by an image. \eat{We use ranking SVM to seek that direction, along which the relevant tags' vector representations rank ahead of the irrelevant tags'.} }
\label{fRank}  
\vspace{-21pt}
\end{figure}

\paragraph{Results.} The MiAP results averaged over all the distinct regulations are reported in Figure~\ref{fRank}(left), in which we test the 300D GloVe vectors~\cite{pennington_glove:_2014} and word2vec~\cite{mikolov_linguistic_2013} of dimensions 100, 300, 500, and 1000. The horizontal axis shows different regularizations we use for training the ranking SVMs. Larger $\lambda$ regularizes the models more. In the 300D GloVe space and the word2vec spaces of 300, 500, and 1000 dimensions, more than two ranking SVMs, with small $\lambda$ values, give rise to nearly perfect ranking results ($\text{MiAP}\approx 1$), showing that the seen tags $\calS$ are linearly rank-able under almost every visual association rule---all the ranking constraints imposed by the relevant $Y_m$ and irrelevant $\irYm$ tags to image $m$ are satisfied. 

However, we shall be cautious before drawing any conclusions \emph{beyond} the experimental vocabulary  $\calS$ of seen tags. An image $m$ incurs a visual association rule essentially over all words, though the same rule implies different partitions  of distinct experimental vocabularies (e.g., the seen tags $\calS$ and unseen ones $\calU$). Accordingly, we would expect the principal direction for the seen tags is also shared by the unseen tags under the same rule, if the answer is YES to the questions at the end of Section~\ref{sStructure}.
\vspace{-15pt}



\paragraph{Generalization to unseen tags.} We test whether the same principal direction exists for  the seen tags and unseen ones  under every visual association rule induced by an image. This can be (only partially) justified by applying the ranking SVMs previously learned, to the unseen tags\rq{} vectors, because we do not  know the ``true'' principal directions. We consider the with 81 unseen tags $\calU$ as the ``test data'' for the trained ranking SVMs, each due to an image incurred visual association. NUS-WIDE provides the annotations of the 81 tags for the images. The results, shown in Figure~\ref{fRank}(right), are significantly better than the most basic baseline, randomly ranking the tags (the black curve close to the origin),  demonstrating that the directions output by SVMs are generalizable to the new vocabulary $\calU$ of words. 
\vspace{-15pt}

\paragraph{Observation.} Therefore, we conclude that the word vectors  are an efficient media to transfer knowledge---the  rank-ability along the principal direction---from the seen tags to the unseen ones. We have empirically verified that the visual association rule $(Y_m,\overline{Y_m})$ in words due to an image $m$ can be represented by the linear rank-ability of the corresponding word vectors along a principal direction. Our experiments involve $|\calS|+|\calU|= $ 1,006 words in total. Larger-scale  and theoretical studies are required for future work.

\eat{
Since we are trying to rank labels represented as embedding vector in a contnous embedding space, let us make a hypothesis: For each unique image, there is a unique corresponding hyperplane in embedding space. In the ideal situation this hyperplane would perfectly rank all tags ,including currently unseen ones, for this specific image in correct order. However we could only approximating such ideal hyperplane from the training tags we have because we never know the occurrence of every possible tag. Since we assume such hyperplanes exist, we should also be able to find a mapping $f(\cdot)$ between image $\vx_{\cst{m}}$ and hyperplane so we could predict such hyperplane directly from given query image $\vx_{\cst{m}_{te}}$.

Hence there are actually two challenges hidden behind the problem: 1, How do we determine such hyperplane from the known incomplete training set. 2, How do we approximate mapping $f(\cdot)$ between an image and a hyperplane. We will discuss how we are trying to solve these two problem in one unified framework in section.~\ref{sApproach}.

But before that, we would like to verify two important presuppositions: 1. Does such ideal linear ranking hyperplane ever exist?  2. It is possible to predict disjoint tags using a hyperplane inferred from seen tags?

\subsection{Rankability and Predictability Sanity Check}

In this section we will justify two above presuppositions: 1. The labels in the embedding space is linearly rankable so that a near-perfect label prediction is achievable in ideal situation. 2. We will be able to transfer ranking information from seen labels to rank unseen labels though embedding space.

For the first perspective, we ran rank-SVM on a validation set of NUS tagging dataset. The validation dataset contains 25689 images as well as combinations of 925 unique labels. There are 21863 unique label combinations among 25698 of them. For each of those label combination, we set the label embeddings of positive tags as positive samples and the rest among 925 unique tags as negative samples. We then train a linear rank hyperplane though rank-SVM on the samples and re-tested them on those samples to observe how linear rankable those word embeddings are. We evaluate the ranking confidence results using MiAP. Please refer to the section.~\ref{sExp} for the computation of MiAP. This test is designed to verify that weather there existing a optimal hyperplane which could perfectly rank the labels in embedding space or not for each common unique label combination. Since 21863 unique combinations of 925  manually annotated labels is not a small number, we believe this test is meaningful.

To better demonstrate the rankability of the embeddings, we use different label embeddings from different methods, including Glove~\cite{pennington_glove:_2014} and Word2vec~\cite{mikolov_linguistic_2013,mikolov_distributed_2013,mikolov_efficient_2013}, in different dimensions. We also apply increasing lambda, which is the trade-off parameter between regularization and ranking loss in rank-SVM.

The Word2vec vectors we used here are trained on the Wikipedia database. And the Glove embedding are the pre-trained model trained on the crawl 840B provided in Glove project homepage. And the results are shown figure in ~\ref{fig:linear_rankable}.

From figure.~\ref{fig:linear_rankable} we could tell that the label embeddings are all generally linear rankable even with high lambda (the hyperplane is highly regularized). This indicates a prefect solution is possible if the mapping $f(\cdot)$ is decent.
}

\eat{
For the second perspective, we tested our hyperplanes trained in the aforementioned experiment on an additional unseen 81 labels annotated on the validation set. The 925 labels in the previous experiment and the 81 labels used here are mutually exclusive. In another word, we are predicting ranks of labels based on some other disjoint labels only. This is similar to tag completion task without using images. This sanity check would reveal how relatively generalized the word embedding could be and how correlated the tags are. And the results are shown figure in ~\ref{fig:linear_generalization}.

We could see from both figures that the linear rankability and predictability are directly correlated to the dimensionality of the word embedding even they are trained on the same text corpus.

We could draw at least two conclusions from figure.~\ref{fig:linear_generalization}. First is that word embedding could be an efficient media for us to transfer knowledge from seen tags to the task of ranking and predicting unseen tags. Second is that although technically they should be independent events, different tags are so statistically correlated that we could predict image tags even without seeing any images or modeling tagging like TagCooccur~\cite{sigurbjornsson_flickr_2008}.
}

\section{Approximating the linear ranking functions}
\label{sApproach}
This section presents our Fast\textbf{0}Tag approach to image tagging. We first describe how to solve image tagging by approximating the principal directions thanks to their existence and generalization, empirically verified in the last section. We then describe  detailed approximation techniques.

\subsection{Image tagging by ranking}

Grounded on the observation from Section~\ref{sRankability}, that there exists a principal direction $\pd_m$, in the word vector space, for every visual association rule $(Y_m,\overline{Y}_m)$ in words by an image $m$, we propose a straightforward solution to image tagging. The main idea is to  approximate the principal direction by learning a mapping function $\vf(\cdot)$, between the visual space and the word vector space, such that 
\begin{align}
\vf(\vx_m)\approx\pd_m,
\end{align}
where $\vx_m$ is the visual feature representation of the image $m$.  Therefore, given a test image $\vx$, we can immediately suggest a list of tags by ranking the word vectors of the tags along the direction $\vf(\vx)$, namely, by  the ranking scores,
\begin{align}
\inner{\vf(\vx)}{{\vt}}, \quad \forall {\vt} \in \calS \cup \calU \label{eRankScore}
\end{align}
no matter the tags are from the seen set $\calS$ or unseen set $\calU$.  

We explore both linear  and nonlinear neural networks for implementing the approximation function $\vf(\vx)\approx\pd$.

\subsection{Approximation by linear regression} \label{sLinear}
Here we assume a  linear function from the input image representation $\vx$ to the output principal direction ${\vw}$, i.e., 
\begin{align}
\vf(\vx) \defeq A{\vx},
\label{eDecisionRule}
\end{align}
where $A$ can be solved in a closed form by linear regression. Accordingly, we have the following from the training 
\begin{align}
{\vw}_m = A{\vx}_m + \vct{\epsilon}_m, m=1,2,\cdots,\cM \label{eRegress}
\end{align}
where ${\vw}_m$ is the principal direction of all offset vectors of the seen tags, for the visual association rule $(Y_m,\irYm)$ due to the image $m$, and $\vct{\epsilon}_m$ are the errors. Minimizing the mean squared errors gives us a closed form solution to $A$.

One caveat is that we do not  know the exact principal directions ${\vw}_m$ at all---the training data only offer images $\vx_m$ and the relevant tags $Y_m$. Here we take the easy alternative and use the directions found by ranking SVMs (cf.\ Section~\ref{sRankability}) in eq.~(\ref{eRegress}). There are thus \emph{two stages} involved to learn the linear function $\vf(\vx) = A{\vx}$. The first stage trains a ranking SVM over the word vectors of seen tags for each visual association $(Y_m,\irYm)$. The second stage solves for the mapping matrix $A$ by linear regression, in which the targets are the directions returned by the ranking SVMs. 
\vspace{-10pt}

\paragraph{Discussion.} We note that the the linear transformation between visual and word vector spaces has been employed before, e.g., for zero-shot classification~\cite{akata_label-embedding_2013,frome_devise:_2013} and image annotation/classification~\cite{weston_wsabie:_2011}. This work differs from them with a prominent feature, that the mapped image $\vf(\vx) = A{\vx}$ has a clear meaning; it depicts the principal direction, which has been empirically verified, for the tags to be assigned to the image. We next extend the linear transformation to a nonlinear one, through a neural network.

\begin{figure}
\centering
\includegraphics[width=0.8\linewidth]{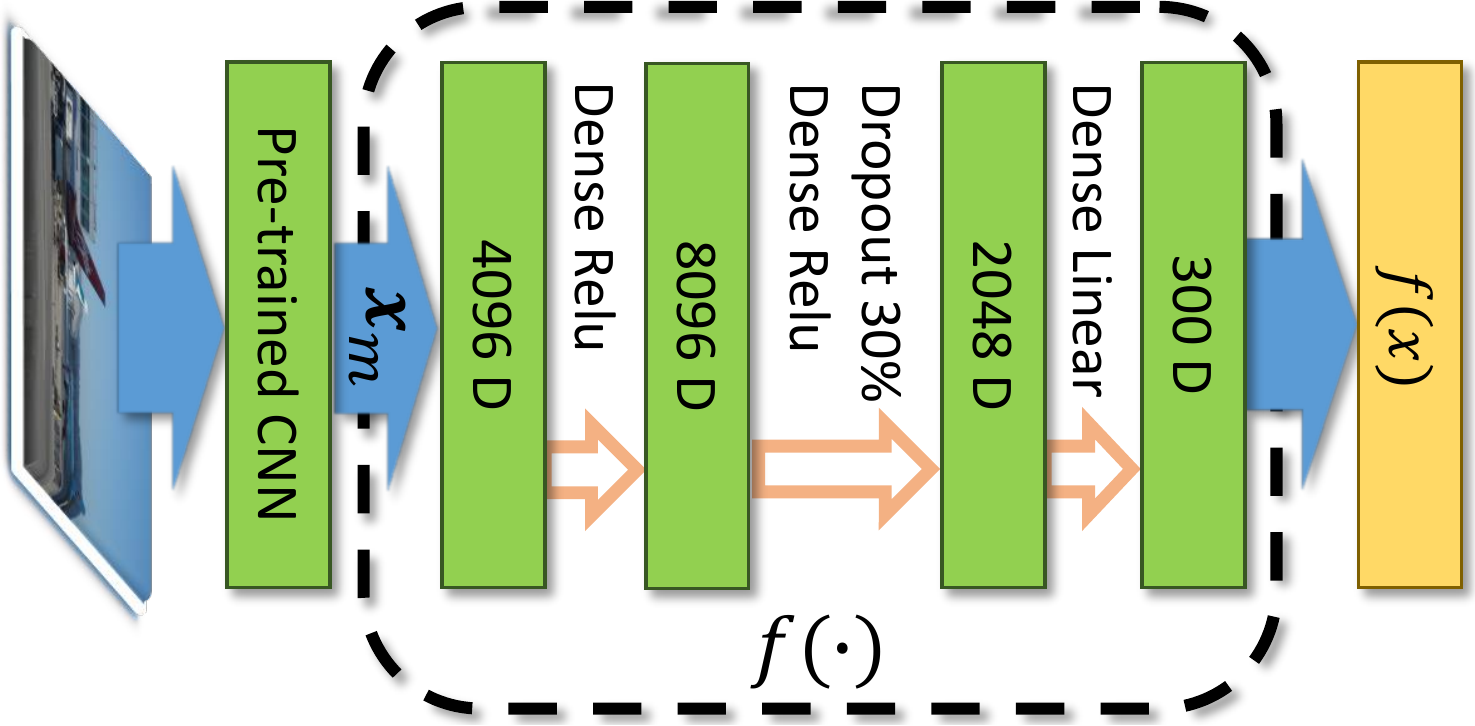}
\caption{The neural network used in our approach for implementing the mapping function $\vf(\vx;\vtheta)$ from the input image, which is represented by the CNN features $\vx$, to its corresponding principal direction in the word vector space.}
\label{fNN}
\vspace{-10pt}
\end{figure}

\subsection{Approximation by neural networks}
We also exploit a nonlinear mapping $\vf(\vx;\vtheta)$ by a multi-layer neural network, where $\vtheta$ denotes the network parameters. Figure~\ref{fNN} shows the network architecture. It consists of two RELU layers followed by a linear layer to output the approximated principal direction, ${\vw}$, for an input image $\vx$. We expect  the nonlinear mapping function $\vf(\vx;\vtheta)$ to offer better modeling flexibility than the linear one. 

Can we still train the neural network  by regressing to the $\cM$ directions obtained from ranking SVMs? Both our intuition and experiments tell that this is a bad idea. The number $\cM$ of training instances is small relative to the number of parameters in the network, making it hard to avoid overfitting. Furthermore, the directions by ranking SVMs are not the true principal directions anyway. There is no reason for us to stick to the ranking SVMs for the principal directions.

We instead unify the two stages in Section~\ref{sLinear}. Recall that we desire the output of the neural network $\vf(\vx_m;\vtheta)$ to be the principal direction, along which all the relevant tag vectors $\vp\in Y_m$ of an image $m$ rank ahead of the irrelevant ones $\vn\in\overline{Y_m}$. Denote by $${\nu}(\vp, \vn;\vtheta)=\inner{\vf(\vx_m;\vtheta)}{\vn} - \inner{\vf(\vx_m;\vtheta)}{\vp},$$ the amount of violation to any of those ranking constraints. We minimize the following  loss to train the neural network,
	\begin{align}
	\vtheta^\star &\leftarrow \argmin_{\vtheta} \quad \sum_{m=1}^{\cM} \omega_m\ell(\vx_m, Y_m;\vtheta),  \label{eNN} \\
	\ell(\vx_m, Y_m;\vtheta) &= \sum_{\vp \in Y_m} \sum_{\vn \in \irYm} \log\left(1+\exp\{\nu(\vp,\vn;\vtheta)\}\right) \notag
	\end{align} 
where $\omega_m=\left(|Y_m||\irYm|\right)^{-1}$ normalizes the per-image RankNet loss~\cite{burges_learning_2005}  $\ell(\vx_m,Y_m;\vtheta)$ by the number of ranking constraints  imposed by the image $m$ over the tags. This formulation enables the function $\vf(\vx)$ to directly  take account of the ranking constraints by relevant $\vp$ and irrelevant $\vn$ tags. Moreover, it can be optimized with no challenge at all by standard mini-batch gradient descent.
\vspace{-11pt}


\paragraph{Practical considerations.}
We use Theano~\cite{bergstra_theano:_2010} to solve the optimization problem. A mini-batch consists of 1,000 images, each of which incurs on average 4,600 pairwise ranking constraints of the tags---we use all pairwise ranking constraints in the optimization. The normalization $\omega_m$ for the per-image ranking loss suppresses the violations from the images with many positive tags. This is desirable since the numbers of relevant tags of the images are unbalanced, ranging from 1 to 20. Without the normalization the MiAP results drop by about 2\% in our experiments. For regularization, we use early stopping and a dropout layer~\cite{hinton_improving_2012} with the drop rate of 30\%. The optimization hyper-parameters are selected by the validation set (cf.\ Section~\ref{sExp}).

In addition to the RankNet loss~\cite{burges_learning_2005} in eq.~(\ref{eNN}), we have also experimented some other choices for the per-image loss, including the hinge loss~\cite{cortes_support-vector_1995}, Crammer-Singer loss~\cite{crammer_algorithmic_2002}, and pairwise max-out ranking~\cite{joachims_optimizing_2002}. The hinge loss performs the worst, likely because it is essentially not designed for ranking problems, though one can still understand it as a point-wise ranking loss. The Crammer-Singer, pairwise max-out, and RankNet are all pair-wise ranking loss functions. They give rise to comparable results and RankNet outperforms the other two by about 2\% in terms of MiAP. This  may attribute to the ease of control over the optimization process for RankNet. Finally, we note that the list-wise ranking loss~\cite{xia_listwise_2008} can also be employed.

\eat{
\subsection{Alternative Methods}
Besides the neural network, we also proposed a series of alternatives to approximated the optimal hyperplane $f(\cdot)$.

\paragraph{Linear Regression.}Despite we consider estimating $f(\cdot)$ as a optimization problem in proposed method, an intuitive alternative would be considering it as a regression problem instead.

First we derive the image feature and rank-SVM hyperplane over seen tags for each image in the train set. Then we solved the regression between the hyperplane weights and image features in th training set in a closed form linear approach. After obtaining the linear mapping matrix between hyperplanes and image feature, we could map the image feature of given query image to its corresponding ranking hyperplane in the semantic space, thus derive its labeling confidence.

\paragraph{Seen2Unssen.}Inspired by the fact that we could predict unseen tags from seen tags of same image as shown in the section~\ref{sRankability}, we developed a baseline which predicts unseen tags from the prediction confidence of seen tags generated in the state-of-art traditional tagging method.

The idea is to train a conventional tagging model on the seen tags training set first. And predict the confidence of seen tags on the testing image. After that we could obtain a unique label ranking list for each image based on their label confidences. Then for each image we train a rank-SVM in the embedding space based the ranking list as we did in section~\ref{sRankability}. Use the rank-SVM in the embedding space, we could obtain the confidence of unseen tags once we know their embeddings.

Please note that this is not a conventional tag completion task since the tag sets are disjoint despite both tasks are very similar.

\paragraph{Linear Fast0Tag.}In the baseline, we are still solving the same optimization problem in the proposed method. But instead we are approximating $f(\cdot)$ using a linear mapping over image feature instead of a neural network.
}

\eat{
In conventional ranking problem, $\mathbf{x}$ are instances represented in continuous space. $\mathbf{y}$ are discrete labels consistent in training set and testing set. A general approach is to estimate a ranking function $f(\vx_m)\rightarrow\vc_m$, where $\vc_m\in\mathbb{R}^{1\times \cN_{tr}}$ quantifies association between the $\cst{m}$th query $\vx_{\cst{m}}$ and $\cN_{tr}$ labels in $\mathbf{y}$.

In this section, we discussed to how we could infer a good linear ranking hyperplane for each image so that the ranking hyperplane could rank both seen and unseen tags.

\subsection{Zero-Shot Ranking Framework}

\begin{table}
	\tiny
\begin{center}
\begin{tabular}{|l|c|}

\hline
{\bf Notations} & {\bf Description} \\
\hline\hline

$\cM_{tr}, \cM_{te}$ & Number of images in training and testing sets\\
$\cN_{tr}, \cN_{te}$ & Number of tags in training and testing sets\\
$\cD_s$ & Dimensionality of semantic word embedding\\
$\cD_v$ & Dimensionality of visual feature\\
$\vct{L}_{tr}\in{\{0,1\}}^{{\cM_{tr}}\times{\cN_{tr}}}$ & Binary labeling matrix for  training set\\
$\vct{L}_{te}\in{\{0,1\}}^{{\cM_{te}}\times{\cN_{te}}}$ & Binary labeling matrix for  testing set\\
$\mat{X}_{tr}\in\mathbb{R}^{{\cN_{tr}}\times{\cD_v}}$ & Image feature matrix for training set\\
$\mat{X}_{te}\in\mathbb{R}^{{\cN_{te}}\times{\cD_v}}$ & Image feature matrix for testing set\\
$\mat{Y}_{tr}\in\mathbb{R}^{{\cN_{tr}}\times{\cD_s}}$ & Label embedding matrix for training set\\
$\mat{Y}_{te}\in\mathbb{R}^{{\cN_{te}}\times{\cD_s}}$ & Label embedding matrix for testing set\\

\hline
\end{tabular}
\end{center}
\caption{Major Notation Summary}
\label{tbl:annotation}
\end{table}

Unlike conventional ranking problem, we are not estimating ranking confidence of discrete $\mathbf{y}$ set directly for each query. Instead for each query $\vx_{\cst{m}_{te}}$ we are mapping it to a unique linear ranking function , represented as a ranking hyperplane weight $\vw_{\cst{m}_{te}}$,in the continuous feature space of $\mathbf{y}$. And that estimated ranking hyperplane should be subjected to a certain ranking loss function during training, so that we can derive the optimal $f(\vx_{\cst{m}_{te}})\rightarrow\vw_{\cst{m}_{te}}$.

In this paper, the function $f(\cdot)$ is approximated using a neural network minimizing loss function
	
	\begin{equation}
	min\sum_{{\cst{m}_{tr}}=1}^{\cst{M}_{tr}}\theta(\mat{Y}_{tr}\times f(\vx_{\cst{m}_{tr}})\, ,\, \vl_{\cst{m}_{tr}})
	\label{eq:rank_loss}
	\end{equation}

where $\vx_{\cst{m}_{tr}}$ is the ${\cst{m}_{tr}}$th image in training set and $\vl_{\cst{m}_{tr}}$ is image's binary labeling vector. Since $f(\vx_{\cst{m}_{tr}})$ is the output of neural network as a hyperplane weight in label embedding space, $\mat{Y}_{tr}\times f(x_{\cst{m}_{tr}})$ will be the labeling confidence on training labeling set. So that $\mat{Y}_{tr}\times f(\vx_{\cst{m}_{tr}})$ would be directly comparable with the label ground-truth $\vl_{\cst{m}_{tr}}$. That enable us to evaluate the loss using a ranking loss function $\theta(\mat{Y}_{tr}\times f(\vx_{\cst{m}_{tr}})\, ,\, \vl_{\cst{m}_{tr}})$. And $\theta(\cdot)$ could be any conventional ranking loss function and we will discuss those we used for our problem in the next section.

There are three major advantage: 1. In this way we could estimate rank of $\vy_{unseen}$ as long as they are homogeneous to $\mathbf{y}$ even they did not present in the train stage. 2. Since each query is represented as a unique linear ranking function in target space, we could quick assign confidence once some new $\vy$s are added to the candidates without re-compute the mapping function.  3. We could incorporate most of the tested ranking loss into our framework during training to infer a better mapping.

\subsection{Optimization Objective}
In our case, we are using weighted pairwise RankNet~\cite{burges_learning_2005} as our rank loss function $\theta(\cdot)$. The pairwise loss function for the $\cst{m}_{tr}$th image during training is 

	\begin{equation}\label{eq:ranknet}
	\begin{split}
	\theta(\mat{Y}_{tr}\times f(\vx_{\cst{m}_{tr}})\, ,\, \vl_{\cst{m}_{tr}})=\sum_{i=1}^{\cst{c}_+}\sum_{j=1}^{\cst{c}_-}\log(1+e^{o_{ij}}) \\
 o_{ij}=(\mat{Y}_{tr}\times f(\vx_{\cst{m}_{tr}}))_j-(\mat{Y}_{tr}\times f(\vx_{\cst{m}_{tr}}))_i
	\end{split}
	\end{equation} 
	
The $(\mat{Y}_{tr}\times f(\vx_{\cst{m}_{tr}}))_i$ is the $i$th dimension of the vector $(\mat{Y}_{tr}\times f(\vx_{\cst{m}_{tr}}))$ quantifying prediction confidence of $i$th labels. And $\cst{c}_+$ and $\cst{c}_-$ are the number of positive and negative tags respectively for the ${\cst{m}_{tr}}$th image. This rank loss function sums the RankNet loss between each positive-negative tag pair in the ${\cst{m}_{tr}}$th image.
	
The final ranking loss function to minimize for a training batch is

\begin{equation}\label{eq:batch_loss}
min\sum_{{\cst{m}_{tr}}=1}^{\cst{b}}\frac{\theta(\mat{Y}_{tr}\times f(\vx_{\cst{m}_{tr}})\, ,\, \vl_{\cst{m}_{tr}})}{\cst{c}_+\cdot\cst{c}_-}
\end{equation} 

Where $\cst{b}$ is the size of one batch of the neural network training set. By minimizing this function, we could obtain the desire $f(\cdot)$ which maps image to a linear ranking hyperplane in the embedding space. We calculate the sub-gradient of equation.~\ref{eq:batch_loss} to update its value during each epoch of training process.

Once we obtained the optimized $f(\cdot)$ after training, we are able to assign label confidence to query images set $\mat{X}_{te}$ by

\begin{equation}\label{eq:testing}
\mat{L}_{te}=\mat{Y}_{te}\times f(\mat{X}_{te})
\end{equation} 

given any testing word vectors $\mat{Y}_{te}$ of labels which you want to rank over the query images.

\begin{figure}
\centering
\includegraphics[width=0.7\linewidth]{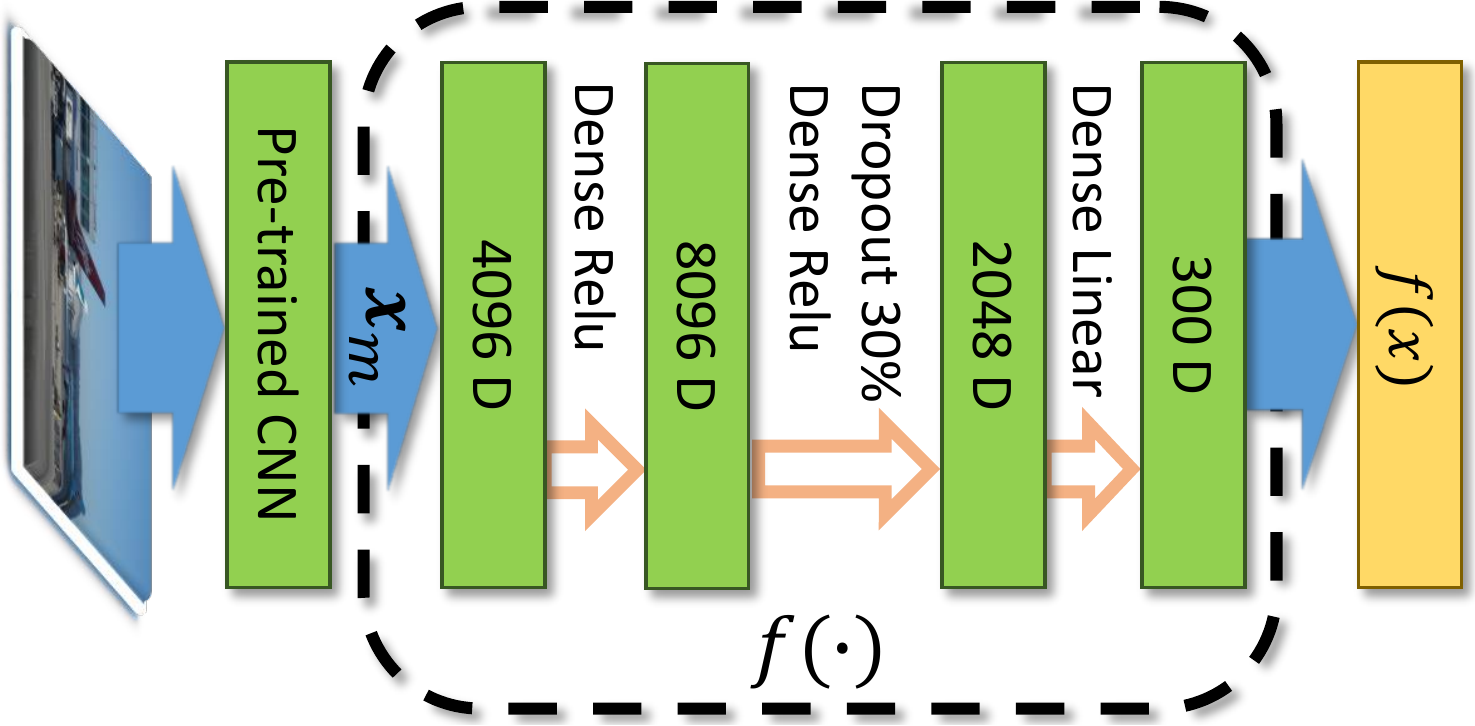}
\caption{This is the visualization of optimizing neural network. Our neural network is a 3-layer fully-connected network following pre-trained CNN. During training, the neural network is optimizing by minimizing $\theta(\cdot)$ which is constrained by two priors: both word vector matrix and label ground-truth. Loss part is no longer needed once training is finished.}
\label{fig:optimization}
\end{figure}

\subsection{Different Ranking Loss}

Our framework could adopt various ranking loss functions such as RankNet~\cite{burges_learning_2005}, Crammer-Singer loss~\cite{crammer_algorithmic_2002}, pairwise max-out ranking~\cite{joachims_optimizing_2002}, list-MLE~\cite{xia_listwise_2008} \etc. Basically they can be categorized in to three classes: Pointwise ranking loss function, which directly model the loss between each label and ground truth. Then second one is the pairwise loss, which focus on the ranking order of pair of labels. The third one is listwise approach which model the ranking between multiple labels. We have investigated how well some pointwise and pairwise ranking algorithms would perform in our framework.

The first one we tested is the hinge loss function~\cite{cortes_support-vector_1995} and square hinge loss function. This is a typical pointwise loss function which solve the labeling problem like a multi-class classification problem. Of course in the task of tagging, one image may belongs to multiple classes simultaneously. We have tested the hinge loss without and with $l2$ regularization term. And the latter one is similar to the SVM objective function. Unfortunately the performance of both loss function is close to random even for traditional tagging problem and very disappointing. In this case we think maybe it is not the optimal loss function for ranking problem.

The second one we are discussing is the Crammer-Singer loss~\cite{crammer_algorithmic_2002} and pairwise max-out ranking~\cite{joachims_optimizing_2002}. They are pairwise ranking loss function which tries to rank positive label higher than negative label in a label pair. These two ranking function are very similar. They only difference is how they chose the pair. These two ranking loss functions really boosts the performance of the model that we are ale to get performance close to the result we are reporting in the experimental section. 

The third one we tested is the RankNet~\cite{burges_learning_2005}. This is also a pairwise loss function. This loss function has improved F1 score and MiAP for 2-3\% compared with the max-out pointwise, and this is the loss function we are using eventually.

We have also went though different strategies of choosing pairs. The most intuitive way is to pick up every possible positive-negative label pair within each image, which is what we are currently adopting in our loss function. We have also tried another strategy, which pick every possible pair within the entire batch instead of just one image. In that case we could use images that has no positive tags as the negative samples during training instead discarding them. But the drawback is also obvious that we will have to use very large extra memory space. It did improve the performance a little, but we think it is not worthy the computational resource it spent,so we keep the original pairing strategy.

We also adopted the pairing weights to balance the loss each image created by adding a $\frac{1}{\cst{c}_+\cdot\cst{c}_-}$ weight in equation.~\ref{eq:batch_loss}. This prevents the model from being overwhelmed towards images that have more label pairs. And this also give us a improvement in MiAP around 2\%.
		
}	
\section{Experiments on NUS-WIDE}
\label{sExp}

This section presents our experimental results. We contrast our approach to several competitive baselines for the conventional image tagging task on the large-scale NUS-WIDE~\cite{chua_nus-wide:_2009} dataset. Moreover, we also evaluate our method on the zero-shot and seen/unseen image tagging problems  (cf.\ Section~\ref{sRegulation}). For the comparison on these problems, we extend some existing zero-shot classification algorithms and consider some variations of our own approach. 

\eat{
There are three major sections of experiments:
1. The traditional tagging problem. We compared our method with conventional tagging methods such as TagProp~\cite{guillaumin_tagprop:_2009} and FastTag~\cite{chen_fast_2013} on the convention tagging task.

2. The zero-shot tagging problem. We compared our method with zero-shot learning methods such as state-of-art Akata \etal~\cite{akata_evaluation_2015} and some of our own proposed baselines on the zero-shot tagging task. In the zero-shot tagging task, the tags in training and testing set are disjoint.

3. The mix tagging problem. In this scenario we trained our method and baselines on the seen tags and tested them on the combination of both seen and unsee tags.
}

\subsection{Dataset and configuration}
\paragraph{NUS-WIDE.}
We mainly use the NUS-WIDE dataset~\cite{chua_nus-wide:_2009} for the experiments in this section. NUS-WIDE is a standard benchmark dataset for image tagging. It contains 269,648 images in the original release and we are able to retrieve 223,821 of them since some images are either corrupted or removed from Flickr. We follow the recommended experiment protocol to split the dataset into a training set with 134,281 images and a test set with 89,603 images. We further randomly separate 20\% from the training set as our validation set for 1) tuning hyper-parameters in our method and the baselines  and 2) conducting the empirical analyses in Section~\ref{sRankability}. 
\vspace{-12pt}

\paragraph{Annotations of NUS-WIDE.}
NUS-WIDE releases three sets of tags associated with  the images. The first set comprises of 81 ``groundtruth'' tags. They are carefully chosen to be representative of the Flickr tags, such as containing both general terms (e.g., $animal$) and specific ones (e.g., $dog$ and $flower$), corresponding to frequent tags on Flickr, etc. Moreover, they are annotated by high-school and college students and are much less noisy than those directly collected from the Web. This 81-tag set is usually taken as the groundtruth for benchmarking different image tagging methods. The second and the third sets of annotations are both harvested from Flickr. There are 1,000 popular Flickr tags in the second set and nearly 5,000 raw tags in the third. 
\vspace{-25pt}


\paragraph{Image features and word vectors.}
We extract and $\ell_2$ normalize the image feature representations of VGG-19~\cite{simonyan_very_2014}. Both GloVe~\cite{pennington_glove:_2014} and Word2vec~\cite{mikolov_linguistic_2013} word vectors are included in our empirical analysis experiments in Section~\ref{sRankability} and the 300D GloVe vectors are used for the remaining experiments. We also $\ell_2$ normalize the word vectors. 
\vspace{-12pt}

\eat{
For the sake of fairness, all the methods and baselines in this paper, including our proposed neural network approach, are using the same image feature set, which is the features extracted from VGG-19 pre-trained CNN~\cite{simonyan_very_2014}. We $l2$ normalized all of them after extraction for all methods.

As for the word embeddings, we have used different word embeddings in different  dimensions as stated in section ~\ref{sRankability}. According to figure ~\ref{fig:linear_generalization}, we choose Glove trained on their common crawl corpus with 300 dimensions because it has much better performance in the sanity check of predicted unseen tags. We also $l2$ normalized word embedding for all the methods which use them. $l2$ normalization generally improve the performance of word vector in different tasks since it equates dot product and cosine similarity~\cite{levy_improving_2015}. 
}

\paragraph{Evaluation.}

We evaluate the tagging results of different methods using two types of metrics. One is the mean image average precision (MiAP), which takes the whole ranking list into account. The other consists of the precision, recall, and F-1 score for the top $K$ tags in the list. We report the results for $K=3$ and $K=5$. Both metrics are commonly used in the previous works on image tagging. We refer the readers to Section 3.3 of~\cite{li_socializing_2015} for how to calculate MiAP and to Section 4.2 of~\cite{gong_deep_2013} for the top-$K$ precision and recall.

\eat{
\begin{table*}
\centering
\caption{Comparison results of the \textbf{conventional} image tagging with 81 tags on NUS-WIDE. }
\label{tConventional}
\small
\begin{tabular}{|l|c|c|c|c|c|c|c|}
\hline
\multirow{2}{*}{ Method~~~~\%} & \multirow{2}{*}{ MiAP} & \multicolumn{3}{c|}{$K=3$} & \multicolumn{3}{|c|}{$K=5$}\\
    & \multirow{1}{*}{} & P & R  & { F1} & P & R & { F1} \\
    
\hline\hline
{ TagProp}~\cite{guillaumin_tagprop:_2009} & 52.6 &  29.3 & 50.0 & 37.0 & 22.0 & 62.5 & 32.5\\
\hline
{ WARP}~\cite{gong_deep_2013} & 47.7 &  26.7 & 45.4 & 33.6 &  20.2 & 57.4 & 29.9\\
\hline
{ FastTag}~\cite{chen_fast_2013} & 40.7 & 22.9 & 39.1 & 28.9 & 19.3 & 54.1 & 28.5\\
\hline
{ Fast\textbf{0}Tag (lin.)}& 52.2 &29.3 & 49.9 & 36.9 & 21.3 & 60.5 & 31.5\\
\hline
{ Fast\textbf{0}Tag (net.)} & {\bf 55.0} &  {\bf 30.7} & {\bf 52.4} & {\bf 38.7} & {\bf 22.7} & {\bf 64.5} & {\bf 33.6}\\
\hline
\end{tabular}
\vspace{-10pt}
\end{table*}
}

\subsection{Conventional image tagging}
Here we report the experiments on the \textbf{conventional}  tagging. The 81 concepts with ``groundtruth''  annotations in NUS-WIDE are used to benchmark  different methods.
\vspace{-11pt} 

\paragraph{Baselines.} We include TagProp~\cite{guillaumin_tagprop:_2009} as the first competitive baseline. It is representative among the nearest neighbor based methods, which in general outperform the parametric methods built from generative models~\cite{barnard_matching_2003,carneiro_supervised_2007}, and gives rise to state-of-the-art results in the experimental study~\cite{li_socializing_2015}. We further compare with  two most recent parametric methods, WARP~\cite{gong_deep_2013} and FastTag~\cite{chen_fast_2013}, both of which are built upon deep architectures though using different models. For a fair comparison, we use the same VGG-19 features for all the methods---the code of TagProp and FastTag is provided by the authors and we implement WARP based on our neural network architecture. Finally, we compare to WSABIE~\cite{weston_wsabie:_2011} and CCA, both correlating images and relevant tags in a low dimensional space. All the hyper-parameters (e.g., the number of nearest neighbors in TagProp and early stopping for WARP) are selected using the validation set. 
\vspace{-13pt}

\begin{table}
\centering
\caption{Comparison results of the \textbf{conventional} image tagging with 81 tags on NUS-WIDE. }
\vspace{-7pt}
\label{tConventional}
\small
\begin{tabular}{|l|c|c|c|c|c|c|c|}
\hline
\multirow{2}{*}{ Method~~~~\%} & \multirow{2}{*}{ MiAP} & \multicolumn{3}{c|}{$K=3$} & \multicolumn{3}{|c|}{$K=5$}\\
    & \multirow{1}{*}{} & P & R  & { F1} & P & R & { F1} \\
    
\hline\hline
{CCA} & 19 &  9 & 15 & 11 & 7 & 20 & 11\\
\hline
{WSABIE}~\cite{weston_wsabie:_2011} & 28 &  16 & 27 & 20 & 12 & 35 & 18\\
\hline
{TagProp}~\cite{guillaumin_tagprop:_2009} & 53 &  29 & 50 & 37 & 22 & 62 & 32\\
\hline
{WARP}~\cite{gong_deep_2013} & 48 &  27 & 45 & 34 &  20 & 57 & 30\\
\hline
{FastTag}~\cite{chen_fast_2013} & 41 & 23 & 39 & 29 & 19 & 54 & 28\\
\hline
{Fast\textbf{0}Tag (lin.)}& 52 &29& 50 & 37 & 21 & 60 & 31\\
\hline
{Fast\textbf{0}Tag (net.)} & {\bf 55} &  {\bf 31} & {\bf 52} & {\bf 39} & {\bf 23} & {\bf 65} & {\bf 34}\\
\hline
\end{tabular}
\vspace{-10pt}
\end{table}

\begin{table*}
    \centering
\caption{Comparison results of the \textbf{zero-shot} and \textbf{seen/unseen} image tagging tasks with 81 unseen tags and 925 seen tags.}
\vspace{-10pt}
\label{tZSL}
\small
\begin{tabular}{|l|c|c|c|c|c|c|c||c|c|c|c|c|c|c|}
\hline
\multirow{3}{*}{ Method~~~~\%} & \multicolumn{7}{c||}{{\bf Zero-shot} image tagging} & \multicolumn{7}{|c|}{\textbf{Seen/unseen} image tagging}\\
\hhline{~--------------} 
& \multirow{2}{*}{ MiAP} & \multicolumn{3}{c|}{$K=3$} & \multicolumn{3}{|c||}{$K=5$} & \multirow{2}{*}{ MiAP} & \multicolumn{3}{c|}{$K=3$} & \multicolumn{3}{|c|}{$K=5$}\\
    & & P & R  & { F1} & P & R & { F1} &  & P & R  & { F1} & P & R & { F1} \\
    
\hline\hline
{\tt Random} & {\tt 7.1}  & {\tt 2.2} & {\tt 3.8} & {\tt 2.8}  & {\tt 2.2} & {\tt 6.1} & {\tt 3.2} & {\tt 1.2}& {\tt 0.6} & {\tt 0.3} & {\tt 0.4} & {\tt 0.6} & {\tt 0.5} & {\tt 0.5}  \\
\hline
Seen2Unseen & 16.7  & 7.3 & 12.5 & 9.2  & 7.0 & 19.7 & 10.3 & 2.8  & 2.1 & 1.1 & 1.4 & 1.9 & 1.6 & 1.8 \\
\hline
 LabelEM~\cite{akata_evaluation_2015} & 23.7 & 11.9 & 20.2 & 14.9  & 10.2 & 28.9 & 15.1 & 8.8 & 8.7 & 4.4 & 5.8  & 7.9 & 6.6 & 7.2 \\
\hline
 LabelEM+~\cite{akata_evaluation_2015} & 24.9 & 12.5 & 21.4 & 15.8 & 10.7 & 30.4 & 15.8 & 10.2  & 11.3 & 5.7 & 7.6  & 9.6 & 8.1 & 8.8 \\
\hline
 ConSE~\cite{norouzi_zero-shot_2013} & 32.4  & 17.7 & 30.1 & 22.3  & 13.7 & 38.8 & 20.2 & 12.5  & 16.7 & 8.4 & 11.2 & 13.5 & 11.3 & 12.3 \\
\hline

{ Fast\textbf{0}Tag (lin.)}& 40.1  & 21.8 & 37.2 & 27.5  & 17.0 & 48.4 & 25.2 & 18.8  & {\bf 22.9} & {\bf 11.5} & {\bf 15.4}  & {\bf 18.7} & {\bf 15.7} & {\bf 17.1} \\
\hline
{ Fast\textbf{0}Tag (net.)} & {\bf 42.2}  & {\bf 22.6} & {\bf 38.4} & {\bf 28.4}  & {\bf 17.6} & {\bf 50.0} & {\bf 26.0} &  {\bf 19.1} & 21.7 & 11.0 & 14.5  & 18.4 & 15.5 & 16.8\\
\hline
{\tt RankSVM } & {\tt 37.0}  & {\tt 19.7} & {\tt 33.3} & {\tt 24.7}  & {\tt 15.2} & {\tt 42.9} & {\tt 22.5} & --& -- & -- & -- & -- & -- & -- \\
\hline
\end{tabular}
\vspace{-5pt}
\end{table*}

\paragraph{Results.} Table~\ref{tConventional} shows the comparison results of TagProp, WARP, FastTag, WSABIE, CCA, and our {\ours} models implemented respectively by the linear mapping and nonlinear neural network. We can see that TagProp performs significantly better than WARP and FastTag. However, TagProp's training and test complexities are very high, being respectively $O(\cM^2)$ and $O(\cM)$ w.r.t.\ the training set size $\cM$. In contrast, both WARP and FastTag are more efficient, with $O(\cM)$ training complexity and constant testing complexity, thanks to their parametric formulation. Our {\ours} with linear mapping gives comparable results to TagProp and {\ours} with the neural network outperforms the other methods. Also,  both implementations  have as low computation complexities as WARP and FastTag.


\subsection{Zero-shot and Seen/Unseen image tagging}
This section presents some results for the two novel image tagging scenarios, \textbf{zero-shot} and \textbf{seen/unseen} tagging. 

\eat{In conventional image tagging, one  assigns tags to images by selecting from a closed and fixed vocabulary of tags. This is rather stringent; in real-world applications new tags show up and become popular over time. \eat{There are about 53M unique tags on Flickr. }How to handle tags previously unseen at training is thus imperative. }

Fu et al.~\cite{fu_transductive_2015} formalized the \textbf{zero-shot} image tagging problem, aiming to annotate test images using a pre-fixed set $\calU$ of unseen tags. Our {\ours} naturally applies to this scenario, by simply ranking the unseen tags with eq.~(\ref{eRankScore}). Furthermore, this paper also considers \textbf{seen/unseen} image tagging which finds both relevant seen tags from $\calS$ and relevant unseen tags from $\calU$ for the test images. The set of unseen tags $\calU$ could be open and dynamically growing. 

In our experiments, we treat the 81 concepts with high-quality user annotations in NUS-WIDE as the unseen set $\calU$ for evaluation and comparison. We use the remaining 925 out of  the 1000 frequent Flickr tags to form the seen set $\calS$---75 tags are shared by the original 81 and 1,000 tags. 
\vspace{-10pt}

\paragraph{Baselines.} Our {\ours} models can be readily applied to the  zero-shot and seen/unseen image tagging scenarios. For comparison we study the following baselines. 
\vspace{-5pt}
\begin{description}    \setlength\itemsep{-1pt}
\item[Seen2Unseen.] We first propose a simple method which  extends an arbitrary traditional image tagging method to also  working with previously unseen tags. It originates from our analysis experiment in Section~\ref{sRankability}. First, we use any existing method to rank the seen tags for a  test image. Second, we train a ranking SVM in the word vector space using the ranking list of the seen tags. Third, we rank unseen (and seen) tags using the learned SVM for zero-shot (and seen/unseen) tagging.
\item[LabelEM.] The label embedding method~\cite{akata_evaluation_2015} achieves impressive results on  zero-shot classification  for fine-grained object recognition. If we consider each tag of $\calS\cup\calU$ as a unique class, though this implies that some classes will have duplicated images, the LabelEM can be directly applied to the two new tagging scenarios. 
\item[LabelEM+.] We also modify the objective loss function of LabelEM when we train the model, by carefully removing the terms that involve duplicated images.  This slightly improves the performance of LabelEM.
\item[ConSE.] Again by considering each tag as a class, we include a recent zero-shot classification method, ConSE~\cite{norouzi_zero-shot_2013} in the following experiments. 
\vspace{-4pt}
\end{description} 
Note that it is computationally infeasible to compare with~\cite{fu_transductive_2015}, which might be the first work to our knowledge on expanding image tagging to handle unseen tags, because it considers all the possible combinations of the unseen tags. 
\vspace{-24pt}

\paragraph{Results.} Table~\ref{tZSL} summarizes the results of the baselines and {\ours} when they are applied to  the zero-shot andseen/unseen image tagging tasks. Overall,  {\ours},  with  either  linear  or neural network mapping, performs the best. 

Additionally, in the table we add two special rows whose results are mainly for reference. The {\tt Random} row corresponds to the case when we return a random list of tags in $\calU$ for zero-shot tagging (and in $\calU\cup\calS$ for seen/unseen tagging) to each test image. We compare this row with the row of Seen2Unseen, in which we extend TagProp to handle the unseen tags. We can see that the results of Unseen2Seen are significantly better than randomly ranking the tags. This tells us that the simple Seen2Unseen is effective in expanding the labeling space of traditional image tagging methods. Some tag completion methods~\cite{sigurbjornsson_flickr_2008} may also be employed for the same purpose as Seen2Unseen.

Another special row in Table~\ref{tZSL} is the last one with {\tt RankSVM} for zero-shot image tagging. We obtain its  results through the following steps. Given a test image, we assume the annotation of the seen tags, $\calS$, are known and then learn a ranking SVM with the default regularization $\lambda=1$. The learned SVM is then used to rank the unseen tags for this image. One may wonder that the results of this row should thus be the upper bound of our {\ours} implemented based on linear regression, because the ranking SVM models are the targets of the linear regresson. However, the results show that they are not. This is not surprising, but rather it reinforces our previous statement that the learned ranking SVMs are not the ``true'' principal directions. The {\ours} implemented by the neural network is an effective alternative for seeking the principal directions. 

It would also be interesting to compare the results in Table~\ref{tZSL} (zero-shot image tagging) with those in Table~\ref{tConventional} (conventional tagging), because the experiments for the two tables share the same testing images and the same candidate tags; they only differ in which tags are used for training. We can see that the {\ours} (net.) results of the zero-shot tagging in Table~\ref{tZSL} are actually comparable to the conventional tagging results in Table~\ref{tConventional}, particularly about the same as FastTag's. These results are encouraging, indicating that it is unnecessary to use all the candidate tags for training in order to have high-quality tagging performance.
\vspace{-10pt}




\begin{table}
	\centering
	\caption{Annotating images with up to 4,093 unseen tags. \vspace{-10pt}}
	\label{t4K}
	\scriptsize
	\begin{tabular}{|l|c|c|c|c|c|c|c|}
		\hline
		\multirow{2}{*}{ Method~~~~\%} & \multirow{2}{*}{ MiAP} & \multicolumn{3}{c|}{$K=3$} & \multicolumn{3}{|c|}{$K=5$}\\
		& \multirow{1}{*}{} & P & R  & { F1} & P & R & { F1} \\
		
			\hline\hline
			{\tt Random} & \tt{0.3}  & \tt{0.1} & \tt{0.1} & \tt{0.1} & \tt{0.1} & \tt{0.1} & \tt{0.1}\\
			\hline
			{ {\ours} (lin.)} & {\bf 9.8}  & {\bf 9.4} & {\bf 7.2} & {\bf 8.2} & {\bf 7.4} & {\bf 9.5} & {\bf 8.4}\\
			\hline
			{ {\ours} (net.)} & 8.5  & 8.0 & 6.2 & 7.0  & 6.5 & 8.3 & 7.3\\
		\hline
	\end{tabular}
	\vspace{-10pt}
\end{table}

\paragraph{Annotating images with 4,093 unseen tags.}
What happens when we have a large number of  unseen tags showing up at the test stage? NUS-WIDE provides noisy annotations for the images with over 5,000 Flickr tags. Excluding the 925 seen tags that are used to train models, there are  4,093 remaining unseen tags. We use the {\ours} models to rank all the unseen tags for the test images and the results are shown in Table~\ref{t4K}. Noting that the noisy annotations weaken the credibility of the evaluation process, the results are reasonably low  but significantly higher than the random lists.
\vspace{-21pt}

\begin{figure}[t]
\centering
\includegraphics[width=\linewidth]{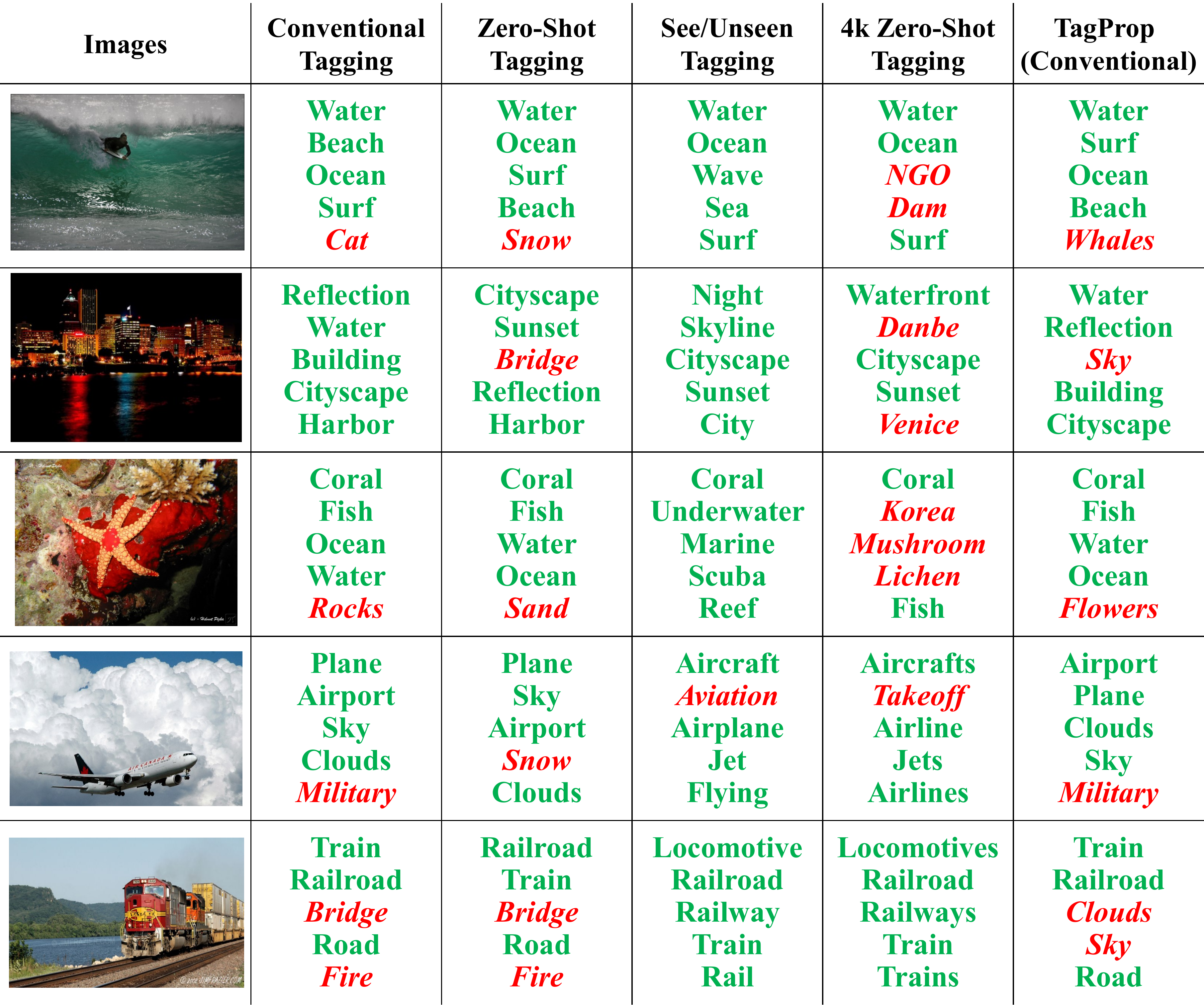}
\vspace{-14pt}
   \caption{The top five tags for  exemplar images~\cite{chua_nus-wide:_2009} returned by  {\ours} on the conventional, zero-shot, and seen/unseen image tagging tasks, and by TagProp for conventional tagging. (Correct tags: {\color{green} green}; mistaken tags: {\color{red} red}  and \textit{italic}. Best viewed in color.)
}
\vspace{-14pt}
\label{fQual}
\end{figure}

\paragraph{Qualitative results.}    Figure~\ref{fQual} shows the top five tags for  some exemplar images~\cite{chua_nus-wide:_2009}, returned by  {\ours} under the conventional, zero-shot, and seen/unseen image tagging scenarios. Those by TagProp under the conventional tagging are shown on the rightmost. The tags in {\color{green} green} color appear in the groundtruth annotation; those in {\color{red} red} color and \textit{italic} font are the mistaken tags. Interestingly, {\ours} performs equally well for traditional and zero-shot tagging and makes even the same mistakes. More results are in Suppl.


\section{Experiments on IAPRTC-12}  \label{sexp}

We present another set of experiments conducted on the widely used  IAPRTC-12~\cite{grubinger_iapr_2006} dataset. We use the same tag annotation and image training-test split as described in~\cite{guillaumin_tagprop:_2009} for our experiments. 

There are 291 unique tags and 19627 images in IAPRTC-12. The dataset is split to 17341  training images and 2286 testing images. We further separate 15\% from the training images as our validation set.

\subsection{Configuration}

Just like the experiments presented in the last section, we evaluate our methods in three different tasks: {\bf conventional} tagging, {\bf zero-shot} tagging, and {\bf seen/unseen} tagging. 

Unlike NUS-WIDE where a relatively small set (81 tags) is considered as the groundtruth annotation, all the 291 tags of IAPRTC-12 are usually used in the previous work to compare different methods. We thus also use all of them conventional tagging. 

As for zero-shot and seen/unseen tagging tasks, we exclude 20\%  from the 291 tags as unseen tags. At the end, we have 233 seen tags and 58 unseen tags.

The visual features, evaluation metrics, word vectors, and baseline methods remain the same as described in the main text.





\begin{table}
\centering
\caption{Comparison results of the \textbf{conventional} image tagging with 291 tags on IAPRTC-12. }
\label{tConventional}
\small
\begin{tabular}{|l|c|c|c|c|c|c|c|}
\hline
\multirow{2}{*}{ Method~~~~\%} & \multirow{2}{*}{ MiAP} & \multicolumn{3}{c|}{$K=3$} & \multicolumn{3}{|c|}{$K=5$}\\
    & \multirow{1}{*}{} & P & R  & { F1} & P & R & { F1} \\
    
\hline\hline
{TagProp}~\cite{guillaumin_tagprop:_2009} & 52 &  54 & 29 & 38 & 46 & 41 & 43\\
\hline
{WARP}~\cite{gong_deep_2013} & 48 &  50 & 27 & 35 &  43 & 38 & 40\\
\hline
{FastTag}~\cite{chen_fast_2013} & 48 & 53 & 28 & 36 & 44 & 39 & 41\\
\hline
{Fast\textbf{0}Tag (lin.)}& 46 & 52 & 28 & 37 & 43 & 38 & 40\\
\hline
{Fast\textbf{0}Tag (net.)} & {\bf 56} &  {\bf 58} & {\bf 31} & {\bf 41} & {\bf 50} & {\bf 44} & {\bf 47}\\
\hline
\end{tabular}
\vspace{-4pt}
\end{table}

\begin{table*}
    \centering
\caption{Comparison results of the \textbf{zero-shot} and \textbf{seen/unseen} image tagging tasks with 58 unseen tags and 233 seen tags.}
\label{tZSL}
\footnotesize
\begin{tabular}{|l|c|c|c|c|c|c|c||c|c|c|c|c|c|c|}
\hline
\multirow{3}{*}{ Method~~~~\%} & \multicolumn{7}{c||}{{\bf Zero-shot} image tagging} & \multicolumn{7}{|c|}{\textbf{Seen/unseen} image tagging}\\
\hhline{~--------------} 
& \multirow{2}{*}{ MiAP} & \multicolumn{3}{c|}{$K=3$} & \multicolumn{3}{|c||}{$K=5$} & \multirow{2}{*}{ MiAP} & \multicolumn{3}{c|}{$K=3$} & \multicolumn{3}{|c|}{$K=5$}\\
    & & P & R  & { F1} & P & R & { F1} &  & P & R  & { F1} & P & R & { F1} \\
    
\hline\hline
{\tt Random} & {\tt 8.1}  & {\tt 2.0} & {\tt 4.5} & {\tt 2.8}  & {\tt 2.2} & {\tt 2.2} & {\tt 8.1} & {\tt 3.5}& {\tt 2.2} & {\tt 1.2} & {\tt 1.5} & {\tt 1.9} & {\tt 1.7} & {\tt 1.8}  \\
\hline
Seen2Unseen & 15.6  & 6.1 & 13.5 & 8.4  & 5.3 & 19.5 & 8.4 & 7.2  & 3.6 & 1.9 & 2.5 & 4.2 & 3.7 & 3.9 \\
\hline
 LabelEM~\cite{akata_evaluation_2015} & 11.5 & 3.6 & 7.9 & 4.9  & 3.6 & 13.3 & 5.7 & 13.8 & 3.1 & 1.7 & 2.2  & 4.4 & 3.9 & 8.7 \\
\hline
 LabelEM+~\cite{akata_evaluation_2015} & 17.6 & 7.3 & 16.1 & 10.0 & 6.4 & 23.4 & 10.0 & 20.1  & 13.9 & 7.4 & 9.7  & 13.2 & 11.8 & 12.5 \\
\hline
 ConSE~\cite{norouzi_zero-shot_2013} & {\bf24.1}  & 9.7 & 21.3 & 13.3  & 8.9 & 32.5 & 13.9 & 32.5  & 38.8 & 20.6 & 26.9 & 31.1 & 27.6 & 29.2 \\
\hline

{ Fast\textbf{0}Tag (lin.)}& 23.1  & {\bf11.3} & {\bf24.9} & {\bf15.6 } & {\bf9.0} & {\bf33.2} & {\bf14.2} & 42.9  & {\bf50.6} & {\bf27.0} & {\bf35.2}  & 40.8 & 36.2 & 38.4 \\
\hline
{ Fast\textbf{0}Tag (net.)} & 20.3  &  8.5 &  18.6 &  11.6  &  7.2 &  26.4 & 11.3 &  {\bf45.9} &  48.2 &  25.7 &  33.5  & {\bf42.2} & {\bf37.4} & {\bf39.7}\\
\hline
{\tt RankSVM } & {\tt 21.6}  & {\tt 10.2} & {\tt 22.6} & {\tt 14.1}  & {\tt 8.6} & {\tt 31.7} & {\tt 13.6} & --& -- & -- & -- & -- & -- & -- \\
\hline
\end{tabular}
\end{table*}

\subsection{Results}

Table~\ref{tConventional}~and~\ref{tZSL} show the results of all the three image tagging scenarios (conventional, zero-shot, and seen/unseen tagging). The proposed {\ours} still outperforms the other competitive baselines in this new IAPRTC-12 dataset.



A notable phenomenon, which is yet less observable on NUS-WIDE probably due to its noisier seen tags, is that the gap between LabelEM+ and LabelEM is significant. It indicates that the traditional zero-shot classification methods are not suitable for either zero-shot or seen/unseen image tagging task. Whereas we can improve the performance by tweaking LabelEM and by carefully removing the terms in its formulation involving the comparison of identical images.

\begin{figure*}
\vspace{4 pt}
\begin{tabular}{cc}
   \includegraphics[height=300pt]{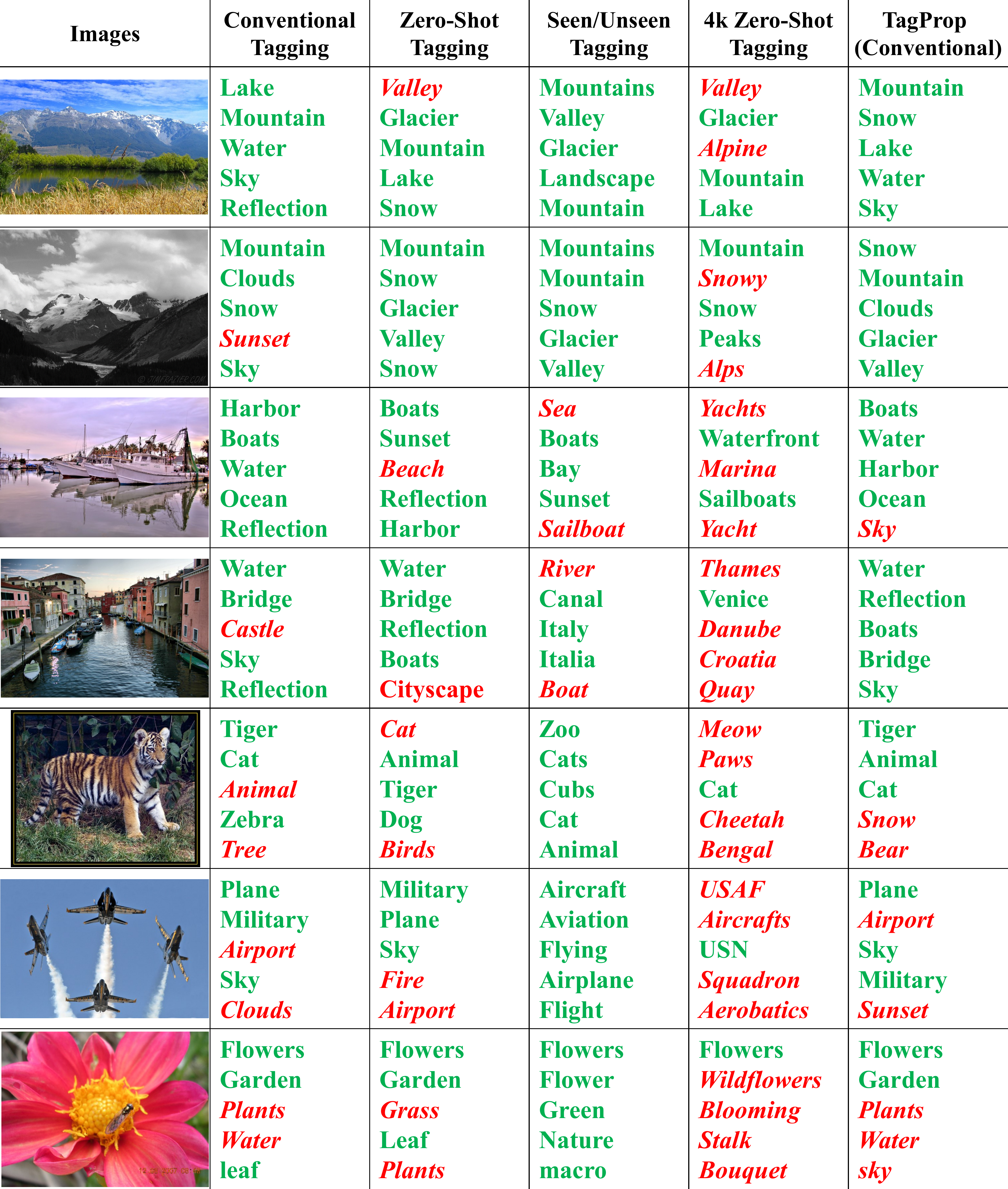} & \includegraphics[height=300pt]{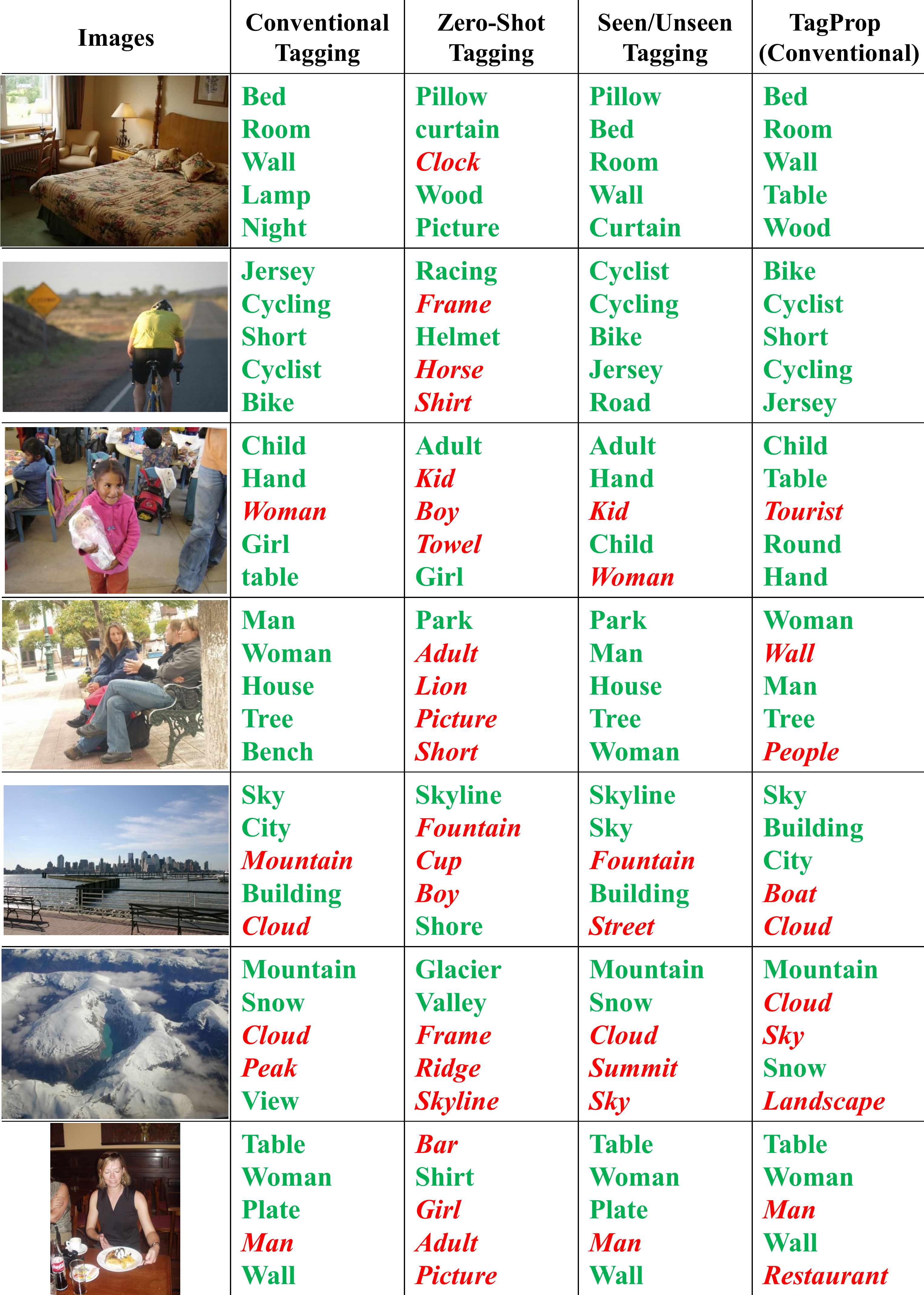}\\
   \vspace{2 pt}\\
   (a) & (b)
\end{tabular}
   \caption{The top five tags for  exemplar images in ~\cite{chua_nus-wide:_2009}(a) and ~\cite{grubinger_iapr_2006}(b) returned by  {\ours} on the conventional, zero-shot, seen/unseen and 4,093 zero-shot image tagging tasks, and by TagProp for conventional tagging. (Correct tags: {\color{green} green}; mistaken tags: {\color{red} red} and \textit{italic})
   }
\label{fQual}  
\end{figure*}

\section{More qualitative results}  \label{squa}

In this section, we provide more qualitative  results of different tagging methods on both the NUS-WIDE, shown in Figure ~\ref{fQual}.(a) supplementing Figure 5 in main text, and the IAPRTC-12, shown in Figure ~\ref{fQual}.(b).

Due to incompletion and noise of tag groundtruth, many actually correct tag predictions are often evaluated as mistaken predictions since they mismatch with groundtruth. This phenomenon becomes especially apparent in 4k zero-shot tagging results in Figure ~\ref{fQual}.(a) where plentiful diverse tag candidates are considered.

\eat{
We evaluate the tagging results by considering both the whole ranking list

Since we are solving a ranking problem, we will compare the performance of difference methods by measuring the rank quality of output labels.

We adopted MiAP (image-centric Mean image Average Precision) from ~\cite{li_socializing_2015} to measure the ranking order. Please refer to section 3.3 in their papers for calculation.

WE also introduced top-$K$ ($K=3,5$ in our experiment) per-class and overall precision and recall score in our experiment as metrics. Please refer to section 4.2 in ~\cite{gong_deep_2013} for the details. Beside these, we calculate the F1 score over the precision recall though harmonic mean.
}

\eat{
\begin{table*}
\tiny
\begin{center}
\begin{tabular}{|l||c|c|c|c|c|c|c|c|c|c|c|c|c|}
\hline
\multirow{3}{*}{\bf Method} & \multirow{3}{*}{\bf MiAP} & \multicolumn{6}{c|}{$k$=3} & \multicolumn{6}{|c}{$k$=5}\\
    \hhline{~~------------} & \multirow{1}{*}{} & \specialcell[c]{per-class\\precision} & \specialcell[c]{per-class\\recall} & \specialcell[c]{\bf per-class\\\bf F1 score} & \specialcell[c]{overall\\precision} & \specialcell[c]{overall\\recall} &\specialcell[c]{\bf overall\\\bf F1 score} & \specialcell[c]{per-class\\precision} & \specialcell[c]{per-class\\recall} & \specialcell[c]{\bf per-class\\\bf F1 score} & \specialcell[c]{overall\\precision} & \specialcell[c]{overall\\recall} &\specialcell[c]{\bf overall\\\bf F1 score} \\

\hline\hline
{\bf WARP}~\cite{gong_deep_2013} & 47.73 & 25.49 & {\bf 47.35} & 30.71 & 26.66 & 45.44 & 33.61 & 18.47 & 47.35 & 26.68 & 20.21 & 57.4 & 29.89\\
\hline
{\bf TagProp}~\cite{guillaumin_tagprop:_2009} & 52.62 & 28.43 & 40.44 & 33.39 & 29.34 & 50.01 & 36.99 & 20.52 & 50.45 & 29.17 & 21.99 & 62.45 & 32.52\\
\hline
{\bf FastTag}~\cite{chen_fast_2013} & 40.66 & {\bf 40.96} & 23.44 & 29.8 & 22.93 & 39.09 & 28.91 & {\bf 33.89} & 36.39 & {\bf 35.09} & 19.25 & 54.09 & 28.48\\
\hline
{\bf Linear Regression}& 52.22 & 27.36 & 41.28 & 32.91 & 29.29 & 49.91 & 36.91 & 18.42 & 50.73 & 27.03 & 21.31 & 60.52 & 31.52\\
\hline
{\bf Fast0Tag} & {\bf 54.97} & 27.83 & 44.74 & {\bf 34.32} & {\bf 30.73} & {\bf 52.37} & {\bf 38.73} & 19.87 & {\bf 54.44} & 29.11 & {\bf 22.72} & {\bf 64.53} & {\bf 33.61}\\
\hline
\end{tabular}
\end{center}
\caption{Traditional Tagging Performance Comparison. This is the evaluation on the NUS test dataset with 81 tags.}
\label{tbl:traditional_result}
\end{table*}
}

\eat{
\subsection{Baseline}

We have introduced multiple state-of-art zero-shot classification as our baseline method. But since they are solving classification problem, they are enforcing similarity between each image and each class in their objective function. So during our comparison, we duplicated the training samples with $n$ multiple labels into $n$ same images with different classes to simulate a multi-class scenario.

\paragraph{Akata+.}Since Akata \etal~\cite{akata_evaluation_2015} has achieved the state-of-art results in the zero-shot learning, it would be an appropriate baseline for us to compare. We also extended their method to multi-label scenario and refer it as Akata+. In their loss function, they are maximizing the confidence difference between the only one positive tag and the maximum negative tag in each image during each iteration. And we are maximizing the confidence difference between the every positive tag and the maximum negative tag in each image during each iteration in order to directly optimize the multi-label problem.

\subsection{Traditional Label Ranking Task}

\subsection{Zero-Shot Label Ranking Task}

In this section, we compared the performance of different methods under different evaluations on the aforementioned zero-shot label ranking configuration. Since  Akata \etal~\cite{akata_evaluation_2015} has achieved the sate-of-art results in zero-shot classification, we will use their method to represent zero-shot classification methods.

\begin{table*}
\tiny
\begin{center}
\begin{tabular}{|l||c|c|c|c|c|c|c|c|c|c|c|c|c|}

\hline
\multirow{3}{*}{\bf Method} & \multirow{3}{*}{\bf MiAP} & \multicolumn{6}{c|}{$k$=3} & \multicolumn{6}{|c}{$k$=5}\\

    \hhline{~~------------} & \multirow{1}{*}{} & \specialcell[c]{per-class\\precision} & \specialcell[c]{per-class\\recall} & \specialcell[c]{\bf per-class\\\bf F1 score} & \specialcell[c]{overall\\precision} & \specialcell[c]{overall\\recall} &\specialcell[c]{\bf overall\\\bf F1 score} & \specialcell[c]{per-class\\precision} & \specialcell[c]{per-class\\recall} & \specialcell[c]{\bf per-class\\\bf F1 score} & \specialcell[c]{overall\\precision} & \specialcell[c]{overall\\recall} &\specialcell[c]{\bf overall\\\bf F1 score} \\

\hline\hline
{\bf Akata \etal}~\cite{akata_evaluation_2015} & 23.67 & 10.12 & 17.28 & 12.77 & 11.85 & 20.2 & 14.94 & 8.61 & 24.27 & 12.71 & 10.17 & 28.89 & 15.05\\
\hline
{\bf Akata+} & 24.93 & 10.75 & 18.75 & 13.66 & 12.54 & 21.36 & 15.8 & 9.08 & 25.94 & 13.45 & 10.68 & 30.33 & 15.79\\
\hline

{\bf ConSE}~\cite{norouzi_zero-shot_2013} & 32.4 & {\bf 27.23} & 20.4 & 23.32 & 17.69 & 30.14 & 22.29 & {\bf 23.35} & 27.17 & {\bf 25.12} & 13.67 & 38.84 & 20.23\\
\hline
{\bf Linear Regression}& 40.13 & 19.32 & 37.44 & 25.49 & 21.84 & 37.22 & 27.53 & 14.74 & 47.79 & 22.53 & 17.04 & 48.4 & 25.21\\
\hline
{\bf Seen2Unseen} & 16.74 & 6.81 & 13.16 & 8.98 & 7.33 & 12.49 & 9.24 & 6.45 & 20.18 & 9.77 & 6.95 & 19.73 & 10.27\\
\hline
{\bf Linear Fast0Tag} & 34.62 & 14.05 & 27.02 & 18.49 & 16.13 & 35.29 & 22.14 & 11.1 & 35.03 & 16.86 & 12.19 & 44.46 & 19.14\\
\hline
{\bf Fast0Tag} & {\bf 42.21} & 21.76 & {\bf 38.04} & {\bf 27.69} & {\bf 22.55} & {\bf 38.43} & {\bf 28.42} & 16.89 & {\bf 48.71} & 25.09 & {\bf 17.6} & {\bf 49.99} & {\bf 26.03}\\
\hline
\end{tabular}
\end{center}
\caption{\vspace{-10pt}Zero-shot Tagging Performance Comparison. This is the evaluation on the NUS test dataset with 81 tags excluded from 925 training tags.}
\label{tbl:ZSL_result}
\end{table*}

We could observe from table.~\ref{tbl:ZSL_result} that our method outperformed the baselines as well as state-of-art methods in the task of zero-shot tagging in all metrics. Although ConSE has slightly outperformed out methods in precision part, but we have near two times higher results in recall. 

\subsection{Mixed Label Ranking Task}

\begin{table*}
\tiny
\begin{center}
\begin{tabular}{|l||c|c|c|c|c|c|c|c|c|c|c|c|c|}

\hline
\multirow{3}{*}{\bf Method} & \multirow{3}{*}{\bf MiAP} & \multicolumn{6}{c|}{$k$=3} & \multicolumn{6}{|c}{$k$=5}\\

    \hhline{~~------------} & \multirow{1}{*}{} & \specialcell[c]{per-class\\precision} & \specialcell[c]{per-class\\recall} & \specialcell[c]{\bf per-class\\\bf F1 score} & \specialcell[c]{overall\\precision} & \specialcell[c]{overall\\recall} &\specialcell[c]{\bf overall\\\bf F1 score} & \specialcell[c]{per-class\\precision} & \specialcell[c]{per-class\\recall} & \specialcell[c]{\bf per-class\\\bf F1 score} & \specialcell[c]{overall\\precision} & \specialcell[c]{overall\\recall} &\specialcell[c]{\bf overall\\\bf F1 score} \\

\hline\hline
{\bf Akata \etal}~\cite{akata_evaluation_2015} & 8.81 & 7.31 & 3.87 & 5.06 & 8.72 & 4.39 & 5.84 & 6.53 & 6.63 & 5.97 & 7.9 & 6.63 & 7.21\\
\hline
{\bf Akata+} & 10.18 & 7.94 & 4.84 & 6.01 & 11.28 & 5.68 & 7.56 & 7.01 & 6.53 & 6.76 & 9.61 & 8.07 & 8.77\\
\hline

{\bf ConSE}~\cite{norouzi_zero-shot_2013} & 12.54 & {\bf 17.98} & 4.05 & 6.61 & 16.7 & 8.41 & 11.19 & {\bf 17.63} & 5.56 & 8.46 & 13.5 & 11.34 & 12.33\\
\hline
{\bf Linear Regression} & 18.78 & 15.37 & {\bf 8.76} & {\bf 11.16} & {\bf 22.91} & {\bf 11.54} & {\bf 15.35} & 13.53 & {\bf 12.32} & {\bf 12.89} & {\bf 18.74} & {\bf 15.74} & {\bf 17.11}\\
\hline
{\bf Seen2Unseen} & 2.82 & 1.81 & 1.02 & 1.31 & 2.14 & 1.08 & 1.43 & 1.68 & 1.55 & 1.61 & 1.93 & 1.62 & 1.76\\
\hline
{\bf Linear Fast0Tag} & 18.29 & 13.35 & 6.23 & 8.49 & 10.92 & 10.92 & 14.52 & 12.37 & 9.34 & 10.64 & 18.42 & 15.47 & 16.81\\
\hline
{\bf Fast0Tag} & {\bf 19.1} & 13.47 & 7.95 & 10 & 21.7 & 10.94 & 14.54 & 12.43 & 11.57 & 11.99 & 18.42 & 15.47 & 16.81\\

\hline
\end{tabular}
\end{center}
\caption{Mixed Tagging Performance Comparison. This is the evaluation on the NUS test dataset with 1006 tags which is the union of 925 training tags and 81 unseen tags.}
\label{tbl:mixed_result}
\end{table*}

\subsection{4K Tags Zero-Shot Label Ranking Task}

\begin{table*}
\tiny
\begin{center}
\begin{tabular}{|l||c|c|c|c|c|c|c|c|c|c|c|c|c|}

\hline
\multirow{3}{*}{\bf Method} & \multirow{3}{*}{\bf MiAP} & \multicolumn{6}{c|}{$k$=3} & \multicolumn{6}{|c}{$k$=5}\\

    \hhline{~~------------} & \multirow{1}{*}{} & \specialcell[c]{per-class\\precision} & \specialcell[c]{per-class\\recall} & \specialcell[c]{\bf per-class\\\bf F1 score} & \specialcell[c]{overall\\precision} & \specialcell[c]{overall\\recall} &\specialcell[c]{\bf overall\\\bf F1 score} & \specialcell[c]{per-class\\precision} & \specialcell[c]{per-class\\recall} & \specialcell[c]{\bf per-class\\\bf F1 score} & \specialcell[c]{overall\\precision} & \specialcell[c]{overall\\recall} &\specialcell[c]{\bf overall\\\bf F1 score} \\

\hline\hline
{\bf ConSE}~\cite{norouzi_zero-shot_2013} & 8.29 & 1.2 & 0.59 & 0.79 & 8.74 & 6.71 & 7.59 & 1.31 & 0.88 & 1.05 & 6.7 & 8.58 & 7.53\\
\hline
{\bf Linear Regression} & {\bf 9.79} & {\bf 2.18} & {\bf 2.34} & {\bf 2.26} & {\bf 9.4} & {\bf 7.22} & {\bf 8.17} & 2.11 & {\bf 3.53} & {\bf 2.64} & {\bf 7.44} & {\bf 9.53} & {\bf 8.36}\\
\hline
{\bf Fast0Tag} & 8.5 & 2.16 & 1.92 & 2.03 & 8.05 & 6.18 & 6.99 & {\bf 2.13} & 2.86 & 2.44 & 6.5 & 8.32 & 7.3\\

\hline
\end{tabular}
\end{center}
\caption{Second Zero-shot Tagging Performance Comparison. This is the evaluation on the NUS test dataset with 4018 tags excluded from 925 training tags.}
\label{tbl:mixed_result}
\end{table*}
}

\section{Conclusion}
\label{sConclusion}

We have systematically studied a particular visual regulation over words, the visual association rule which partitions words into two disjoint sets according to their relevances to an image, as well as how it can be captured by the  vector offsets in the word vector space. Our empirical results show that, for any image, there exists a principal direction in the word vector space such that the relevant tags\rq{}  vectors rank ahead of the irrelevant ones\rq{} along that direction. The experimental analyses involve 1,006 words;  larger-scale and theoretical analyses are required for future work. Built upon this observation, we develop a {\ours} model to solve image tagging by estimating the principal directions for input images. Our approach is as efficient as FastTag~\cite{chen_fast_2013} and is capable of annotating images with a large number of previously {\bf unseen} tags. Extensive experiments validate the effectiveness of our {\ours} approach.

\section*{Acknowledgments} This work is partially supported by NSF IIS 1566511. We thank the anonymous area chair and reviewers, especially the assigned Reviewer 30, for their helpful comments.

\end{document}